\documentclass[3p, preprint, 12pt]{elsarticle}

\usepackage{graphicx}
\usepackage{amssymb}
\usepackage{lineno}
\usepackage{threeparttable}
\usepackage{amsmath}
\usepackage{mathtools}
\usepackage{bm}
\usepackage{subcaption}
\usepackage{booktabs}
\usepackage{hyperref}
\usepackage{multirow}
\usepackage{xcolor}
\usepackage[ruled,vlined]{algorithm2e}
\usepackage{enumitem}
\usepackage{makecell}
\usepackage{xcolor}
\usepackage{import}
\usepackage{graphicx,url}
\usepackage{graphics}
\usepackage{tabularx} % To control the table width
\usepackage{pdflscape}  % Load pdflscape after geometry
\usepackage{afterpage}

\begin{document}

\begin{frontmatter}

%% Title, authors and addresses

%
\title{Analyzing sequential activity and travel decisions with interpretable deep inverse reinforcement learning}
% Large Language Models as Demand Predictors in Event Areas
%\title{Using Textual Data and Large Language Models to Enhance \\Travel Demand Prediction under Planned Special Events}
%Travel Demand

\author[a]{Yuebing Liang} 
\author[b]{Shenhao Wang\corref{cor1}}
\author[c]{Jiangbo Yu}
\author[d]{Zhan Zhao}
\author[e]{Jinhua Zhao}
\author[f]{Sandy Pentland}

\address[a]{The Singapore-MIT Alliance for Research and Technology}
\address[b]{Department of Urban and Regional Planning, University of Florida}
\address[c]{Department of Civil Engineering, McGill University}
\address[d]{Department of Urban Planning and Design, The University of Hong Kong}
\address[e]{Department of Urban Studies and Planning, Massachusetts Institute of Technology}
\address[f]{Media Lab,  Institute for Data, Systems, and Society, Massachusetts Institute of Technology}

\cortext[cor1]{Corresponding author (shenhaowang@ufl.edu)}

\begin{abstract}
Travel demand modeling has shifted from aggregated trip-based models to behavior-oriented activity-based models because daily trips are essentially driven by human activities. To analyze the sequential activity-travel decisions, deep inverse reinforcement learning (DIRL) has proven effective in learning the decision mechanisms by approximating a reward function to represent preferences and a policy function to replicate observed behavior using deep neural networks (DNNs). However, most existing research has focused on using DIRL to enhance only prediction accuracy, with limited exploration into interpreting the underlying decision mechanisms guiding sequential decision-making. To address this gap, we introduce an interpretable DIRL framework for analyzing activity-travel decision processes, bridging the gap between data-driven machine learning and theory-driven behavioral models. Our proposed framework adapts an adversarial IRL approach to infer the reward and policy functions of activity-travel behavior. The policy function is interpreted through a surrogate interpretable model based on choice probabilities from the policy function, while the reward function is interpreted by deriving both short-term rewards and long-term returns for various activity-travel patterns. Our analysis of real-world travel survey data reveals promising results in two key areas: (i) behavioral pattern insights from the policy function, highlighting critical factors in decision-making and variations among socio-demographic groups, and (ii) behavioral preference insights from the reward function, indicating the utility individuals gain from specific activity sequences. 
\end{abstract}

\begin{keyword}
Inverse Reinforcement Learning \sep Explainable Artificial Intelligence \sep Deep Learning \sep Activity-Based Travel Demand Model
%% keywords here, in the form: keyword \sep keyword

%% MSC codes here, in the form: \MSC code \sep code
%% or \MSC[2008] code \sep code (2000 is the default)

\end{keyword}

\end{frontmatter}

%%
%% Start line numbering here if you want
%%
%\linenumbers

%% main text
\section{Introduction} \label{sec:intro}
%  Different from emergency events, PSEs  and
In recent decades, there has been increasing interest in behaviorally-oriented activity-based travel demand models. Unlike simple aggregated trip-based models \citep{mcnally2007four}, which focus solely on individual trips, activity-based models consider the entire sequence of activities individuals undertake throughout the day, such as work, school, shopping, recreation, and other activities \citep{bhat1999activity,mcnally2007activity}. By recognizing the interactions between activities and travel, these models provide deeper insights into how people organize their daily lives and make travel decisions. Understanding these patterns is essential for urban planners and transportation engineers, as it allows them to evaluate the impacts of policy interventions and transportation investments on human behavior \citep{miller2003prototype,arentze2000albatross}.

Early studies on modeling activity and travel decisions utilized rule-based computational process models \citep{miller2003prototype, arentze2000albatross}, which relied on predefined decision rules to simulate human decision-making, often suffered from subjective biases. 
Subsequent research introduced utility-based econometric models, which estimate utility functions for different alternatives and assume that individuals select the alternative with the highest utility. Recent advancements have further enhanced this approach by developing dynamic utility-based models to capture sequential decision-making processes \citep{zimmermann2018capturing, vastberg2020dynamic, song2022modeling}. These models incorporate forward-looking behavior, enabling individuals to consider the future consequences of their decisions and optimize their overall utility. Despite these improvements, existing dynamic utility-based models often depend on discrete choice modeling frameworks that assume simple linear-in-parameter utility functions. Although there have been attempts to enhance the flexibility of utility functions using multiplicative interactions among attributes or mixtures of distributions \citep{tinessa2021closed}, these approaches may still fall short in accurately reflecting behavioral realism.

Recently, machine learning has emerged as a powerful tool for capturing complex patterns in human behavior \citep{drchal2019data, phan2021novel,liang2021nettraj}. Among these techniques, inverse reinforcement learning (IRL) stands out as a specialized approach for modeling sequential decision-making tasks \citep{ng_algorithms_2000,abbeel_apprenticeship_2004}. Assuming observed human behavior is near-optimal with respect to some unknown reward function, the goal of IRL is to infer this underlying reward function, representing the agent's preferences or objectives, and to derive a policy function that replicates the observed behavior. Similar to the utility in microeconometric models, the reward in IRL is a concise and portable representation of a task \citep{song2024state}. Recent research has theoretically demonstrated that IRL and DDCM, though originating from different research fields, belong to a class of optimization problems characterized by a common objective form \citep{sanghvi2021inverse}. Consequently, IRL serves as a bridge between data-driven machine learning and theory-driven behavioral modeling, offering significant potential for interpreting the underlying mechanism that guide human sequential decision-making \citep{alsaleh2020modeling,liu2022understanding,ye2024safety}. Recent advancements have integrated deep neural networks (DNNs) to approximate the reward and policy functions in the IRL framework, enhancing IRL’s ability to capture complex behavioral patterns. This approach, known as deep IRL (DIRL), has been applied to model various sequential decision-making processes in transportation, such as vehicle trajectory generation \citep{choi2021trajgail}, route choice modeling \citep{zhao2023deep}, and activity-travel planning \citep{song2024state}. However, existing studies primarily focus on leveraging DIRL to improve prediction accuracy, often overlooking its potential for interpreting why humans make particular decisions.

This study addresses the research gap by introducing an interpretable DIRL framework for analyzing sequential activity-travel decisions. We model the decision-making process as a Markov Decision Process (MDP) and adapt an adversarial IRL framework \citep{fu_learning_2018} to infer reward and policy functions. The adversarial IRL framework combines IRL with generative adversarial networks (GANs), consisting of a generator to create realistic trajectories and a discriminator to distinguish between real and generated trajectories. The generator estimates a policy function representing human behavioral patterns, while the discriminator estimates a reward function reflecting human behavioral preferences. 
Utilizing the well-trained adversarial IRL model, we derive insights into behavioral patterns from the policy function and behavioral preferences from the reward function. %using a knowledge distillation method \citep{hinton2015distilling} and the reward function using an example-based analysis method \citep{poche2023natural}. 
Experiments with travel survey data demonstrate DIRL's effectiveness in both improving prediction accuracy and enhancing the understanding of activity-travel behavior patterns and preferences. In summary, this study's contributions lie in the following aspects:
% example-based analysis method
% and the reward function using an example-based analysis method \citep{poche2023natural}. 
\begin{itemize}
    \item Unlike previous studies that focus on prediction accuracy, this research introduces an interpretable DIRL framework for analyzing sequential activity and travel decisions. We first train an adversarial IRL model to learn activity-travel behavioral patterns and then introduce a post-hoc interpretation framework to explain the knowledge learned by the policy and reward networks. This interpretation framework is general and can be applied to other sequential human behaviors.
    \item To extract behavioral patterns from the policy function, we adapt a knowledge distillation method, training a surrogate, interpretable model based on the soft labels predicted by a policy network. To explain behavioral preferences from the reward network, we map discrete human behavior sequences to continuous reward values using the learned reward function. These reward values are then used to identify different types of decision-makers and to compare the preference of different activity sequences.
    %use an example-based analysis method, estimating both immediate and accumulated expected utilities based on observed behavior data.
    \item We conducted an analysis using travel survey data from Singapore. The results demonstrate DIRL's interpretability in two key aspects: (i) insights into behavioral patterns from the policy function, highlighting critical factors in decision-making and variations among socio-demographic groups, and (ii) insights into behavioral preferences from the reward function, providing a direct quantifiable measure of utility when making sequential decisions.
\end{itemize}

\section{Literature Review}
\label{sec:theory}

\subsection{Activity and travel decision modeling}

Existing methods for modeling activity and travel decisions can be grouped into three categories: rule-based, utility-based, and machine learning-based approaches. Rule-based models use pre-defined sets of condition-action (if-then) rules to outline how individuals make decisions. Examples include ALBATROSS \citep{arentze2000albatross}, TASHA \citep{miller2003prototype}, and ADAPTS \citep{auld2009framework}. These models often require detailed knowledge of travel patterns and can be biased by subjective interpretations, which may affect their general applicability.

Utility-based models operate on the principle that individuals select activity patterns that maximize their personal utility. These models often break down activity-travel patterns into hierarchical structures to represent various aspects of activity choices. For example, a nested logit structure developed by \cite{bowman2001activity} represents activity patterns at a higher level and details regarding primary and secondary tours at a lower level. %Similarly, \cite{bhat2004comprehensive} proposed a framework that sequentially identifies activities, then builds a complete pattern of an individual's activity and travel. 
\cite{pendyala2005florida} decomposed activity-travel simulation into models for activity type and activity duration, using nested logit and split-population survival models respectively. However, these models are criticized for their inability to comprehensively capture the large space of possible activity-travel patterns, often missing out on the temporal aspects of trips and the interconnected nature of different decision components \citep{vastberg2020dynamic}. Recent advancements have explored dynamic discrete choice theories to improve the modeling of activity-travel decisions. These models assume that individuals make activity-travel decisions sequentially and consider not only the immediate utility of choices but also the anticipated future utility at each decision step. \cite{zimmermann2018capturing} developed a mixed recursive logit model that treats activity-travel scheduling decisions as choices of paths. \cite{vastberg2020dynamic} introduced a dynamic discrete choice model (DDCM) for daily activity-travel planning, which allows agents to be forward-looking by maximizing the anticipated utilities of all travel and activity episodes for the remainder of the day. \cite{song2022modeling} further introduced a mixed dynamic discrete choice model with finite mixture distributions to account for persistent unobserved differences among individuals. An important advantage of utility-based models is that they provide a quantifiable measure of preference or benefit through utility functions, which allows for straightforward interpretations of how different attributes impact human decisions. However, %these models are based on theoretical assumptions like utility maximization and bounded rationality, which can bias their application in real-world scenarios \citep{kahneman2006anomalies}. Additionally, 
designing utility functions is challenging due to the complex nature of activity-travel patterns, and simple linear approaches often fail to capture their intricacies \citep{song2024state}.

Another group of studies has introduced machine learning models in the problem of activity-travel pattern modeling, including decision trees \citep{drchal2019data}, multilayer perceptron networks \citep{phan2021novel}, and random forest \citep{nayak2023joint}. Compared with rule-based and utility-based models, machine learning models are better at capturing complex behavior patterns from large-scale human mobility data and thus can improve the prediction accuracy. On the other hand, their interpretability regarding behavioral mechanisms has been criticized due to their data-driven nature. Recently, inverse reinforcement learning (IRL) has emerged as a powerful paradigm to model sequential decision processes that bridges the gap between data-driven machine learning models and theory-driven behavioral modeling \citep{ng_algorithms_2000, abbeel_apprenticeship_2004}. It assumes that individuals make action decisions to maximize accumulative reward in the future and extract a reward function from demonstration data that explains observed human behavior \citep{abbeel_apprenticeship_2004,ng_algorithms_2000}. The reward function is similar to the utility function in dynamic utility-based models. In fact, recent research has demonstrated deeper connections between dynamic DCMs and IRL: they belong to a class of optimization problems with the same objective and policy definitions, while the main difference is how they approximate the value function for solving the optimization problem \citep{sanghvi2021inverse}. Therefore, compared with other machine learning models, IRL allows for better interpretations regarding underlying behavioral mechanisms. 

Increasing studies have introduced IRL to model sequential decisions in transportation tasks. \cite{ziebart2008maximum} proposed the maximum entropy IRL (MaxEntIRL) for route choice modeling, which assumes linear reward/utility functions based on the principle of maximum entropy to resolve the ambiguity in choosing a distribution over decisions. Subsequent research leveraged learned reward function weights to interpret decision-making mechanisms in travel behavior, including mixed traffic interactions \citep{alsaleh2020modeling}, departure time choices \citep{liu_personalized_2022}, and pedestrian crossing behaviors \citep{ye2024safety}. To extend IRL to more complex and large-scale problems, researchers have incorporated deep learning techniques, leading to the development of Deep IRL (DIRL). For instance, \cite{liu_integrating_2020} used deep learning architectures for reward function approximation in the MaxEntIRL approach in the context of delivery route planning.
More recent work has introduced Adversarial IRL (AIRL), which uses a Generative Adversarial Network (GAN) to solve IRL problems \citep{fu_learning_2018}. %In this way, the model's computation cost for value iteration can be greatly reduced, while an unshaped reward function can be recovered. 
\cite{zhao2023deep} adapted the AIRL framework for route choice modeling with context-dependent reward functions to explicitly consider destination-related factors. \cite{song2024state} refined the AIRL framework to minimize f-divergences between expert and agent state marginal distributions, which were then used to model activity and travel choices. Despite these advancements, most research has predominantly focused on the prediction capabilities of DIRL to improve modeling accuracy, with limited exploration into its interpretability for providing insights into human decision-making strategies.

\subsection {Interpretable Deep Learning}

\noindent Although deep learning models often achieve better prediction performance than their conventional theory-driven statistical or econometric counterparts, their data-driven nature makes it challenging to explain the reasons behind their predictions. This lack of interpretability can raise concerns about the trustworthiness of deep learning models \citep{alwosheel2021did}. Additionally, in applications such as travel behavior modeling, understanding the underlying mechanisms of human decision-making processes is crucial. These insights are valuable for urban designers and policymakers when creating policy interventions or transportation investments. Consequently, there has been increasing attention on enhancing the interpretability of deep learning models \citep{cantarella2005multilayer}. Notable examples include \cite{hinton2015distilling}, who introduced a global surrogate method that distills knowledge from DNNs by re-training interpretable models to fit the predicted soft labels of DNNs. \cite{ribeiro2016should} proposed a local surrogate model called LIME, which uses an interpretable model to fit machine learning model predictions locally. Building on this, \cite{lundberg2017unified} introduced SHapley Additive exPlanations (SHAP), a game theory-based method that connects LIME with Shapley values. Recent research efforts have also focused on interpreting specific types of DNNs, such as Grad-CAM, which provides visual explanations for CNN decisions \citep{ribeiro2016should}, and GNNExplainer, which interprets graph-based machine learning models \citep{ying2019gnnexplainer}.

In the field of transportation, existing studies have linked DNNs with traditional economic models like DCMs and compared their interpretability, although most efforts have focused on mode choice. \cite{wang2020deep} used DNNs for choice analysis and demonstrated that DNNs can extract complete economic information for interpretation, similar to DCMs, including choice predictions, social welfare, elasticities, and more. \cite{sifringer2020enhancing} augmented the utility specification of DCMs with a new nonlinear representation from DNNs, allowing for more flexible utility representations. Researchers also designed new deep learning architectures to integrate the behavioral knowledge into standard feed-forward neural networks, achieving higher predictive performance \citep{Wang2020_dnn_asu, Wang2021_dnn_stat_learning}. \cite{alwosheel2021did} introduced a heat map-based method to explain DNN predictions for mode choice using Layer-wise Relevance Propagation. These studies highlight the close relationship between multi-layer perceptrons (MLPs) and DCMs and provide examples of bridging the gap between data-driven machine learning models and theory-based econometric models using statistical learning theory \citep{Wang2021_dnn_stat_learning}. However, unlike mode choice, the activity-travel decision process is more complex because it is sequential, with individuals making decisions based not only on immediate rewards but also on future expected utilities. Despite the great potential of DIRL in interpreting such sequential decision processes, as discussed in the previous section, existing DIRL-based studies have focused more on prediction accuracy, with limited exploration of its interpretability \citep{zhao2023deep, song2024state}.
%To capture such time-dependent choice behavior, economic researchers have proposed dynamic DCMs, while machine learning researchers have proposed inverse reinforcement learning (IRL). Both approaches estimate a reward/utility function, providing a quantifiable measure of human preferences, and share the common assumption that individuals make decisions at each time step to maximize future expected cumulative reward/utility. Although developed in distinct research fields, recent research has validated their deeper theoretical connections \citep{sanghvi2021inverse}. This equips IRL with unique advantages, as it can improve prediction accuracy from a machine learning perspective and explain behavioral mechanisms through the reward function from an econometric perspective. However, as highlighted in the previous section, existing studies using IRL for travel behavior modeling have predominantly focused on prediction accuracy, with little exploration into how to explain the knowledge that IRL models have learned from humans \citep{zhao2023deep, song2024state}.

% (1) Sequential route decisions are the predominant applications [X,X,X,X]. 

% (2) Computer science literature presents various approaches to address the computational challenges in DIRL. 

% (3) Utility interpretation for DL and travel behavioral modeling. 

% \begin{flalign} \label{eq:choice_probability_dnn}
% s(x_i, w) = \sigma(\Phi_1(x_i, w)) = \frac{1}{1 + e^{-\Phi_1(x_i, w)}}
% \end{flalign}

\section{Methodology}
\label{sec:method}

\begin{figure}[!ht]
  \centering
  \includegraphics[width=\textwidth]{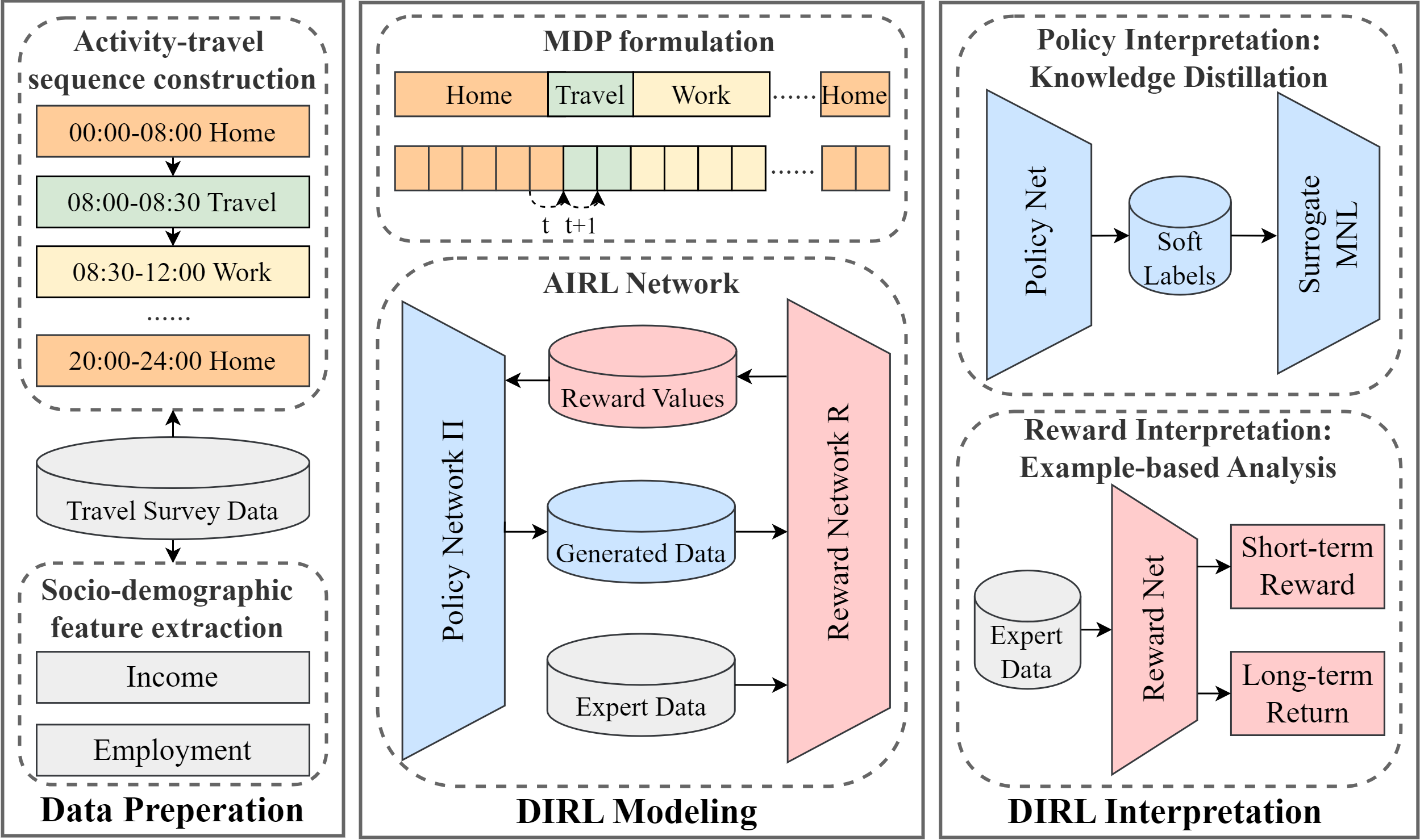}
  \caption{The framework of interpretable DIRL for activity-travel choice modeling}\label{fig:interpretable_DIRL}
\end{figure}

\noindent In this section, we introduce an interpretable DIRL framework to learn and analyze the sequential activity-travel choice process. As shown in Figure~\ref{fig:interpretable_DIRL}, the framework consists of three main components: (1) Data Preperation, extracting activity-travel sequences and socio-demographic features from observed data. This study constructs these features from travel survey data, as will be detailed in Section~\ref{exp:data}. (2) DIRL Modeling, formulating activity-travel decisions as a Markoc Decision Process (MDP) and training an AIRL model to mimic observed behavior. (3) DIRL Interpretation, which consists of a policy interpretator to understand behavioral patterns and a reward interpretator to understand behavioral preferences. More details regarding DIRL modeling and DIRL interpretation will be provided in Section~\ref{method: modeling} and Section~\ref{method: interpretation}, respectively.

%An AIRL\citep{fu_learning_2018} model consists of two DNNs, i.e., a policy net (learning a non-linear policy function) and a reward net (learning a non-linear reward function). To explain the behavioral mechanisms learned by AIRL, we then propose an interpretale DIRL framework, which consists of a policy interpretator and a reward interpretator. The policy interpretator is based on a knowledge distillation method, which uses a surrogate interpretable model to mimic the policy network predictions. The reward interpretator uses example-based explanations to quantify the instant and long-term utility people expected when making activity-trip decisons. (Figure~)

\subsection{DIRL modeling}\label{method: modeling}

In this subsection, we first formulate activity-travel decision-making process as a Markov Decision Process (see Section~\ref{model:MDP}) and then introduce an adapted AIRL model to learn the behavioral mechanisms (see Section~\ref{model:AIRL}).

\subsubsection{Sequential activity-travel choice as MDPs}\label{model:MDP}

This study aims to model daily activity-travel decisions of individuals, focusing on what activities people choose throughout the day and when they decide to start and end these activities or travel behaviors. To represent time choices discretely, we divide the day into fixed-length intervals (e.g., 15 minutes). In this way, an individual's daily activity-travel sequence $Y_m$ can be represented as $Y_m=\{y_1, y_2,...,y_n\}$, where $y_i$ denotes the activity conducted at time step $i$, and $n$ is the total number of time steps in a day.

Markov decision processes (MDPs) provide a mathematical framework for modeling decision-making processes. An MDP is generally defined as $M=\{S,A,T,R, \gamma\}$, where $S$ is the state space, $A$ is the action space, $T(s,a,s')$ is the transition model determining the next state $s'$ given the current state $s$ and action $a$, $R$ is the reward function for each state-action pair, indicating the instantaneous reward similar to the utility function in DDCM models, and $\gamma$ is the discount factor. An agent in state $s \in S$ decides the action $a \in A$ based on a policy $\pi$, which specifies a probability distribution over actions. In the context of activity and travel behavior modeling, each MDP component is defined as follows:
\begin{itemize}
    \item State $s \in S$: The state consists of five travel-related features at the current time step ($Card(S)=5$): (1) Current time step $i \in \{1,2,...,N\}$, (2) Current activity $y_i$, (3) Duration of stay at the current activity $d_i \in \{1,2...N\}$, (4) Number of activities conducted since the start of the day $p_i \in \{1,2...N\}$ (5) Duration since last leaving home $q_i$, 0 if currently at home, otherwise $q_i \in \{1,2...N\}$.
    \item Action $a \in A$: An action represents the activity the agent plans to conduct next, which can include home, work, education, shopping, etc. To explicitly account for travel behavior, travel is defined as an additional activity category. Unlike other activities, travel serves as a transition action when people decide to start a new activity. To represent this, we add action masks in our model to restrict the feasible action space based on the current activity. For example, when the user is engaged in an activity, the action space allows for continuing the same activity or transitioning to travel. When the user is currently traveling, the action space includes travel or any other activity. %Additionally, we incorporate a binary feature to indicate whether the user continues with the same activity or transitions to a different one.
    \item Transition Model $T(s,a,s')$: For activity-travel behavior analysis, the transition model is deterministic, meaning a given state-action pair will always lead to the same next state. Specifically, at the next time step $i+1$, the activity is determined by the action ($y_{i+1}=a_i$). If the agent continues the same activity ($y_{i+1} = y_{i}$), the stay duration at the activity increments by one ($d_{i+1}=d_i + 1$), and the number of activities since the start of the day remains unchanged ($p_{i+1}=p_i$).  If the agent transitions to a different activity, the stay duration resets to one ($d_{i+1}=1$), and the number of activities increases by one ($p_{i+1}=p_i+1$).  The duration since last leaving home increments by one ($q_{i+1}=q_i + 1$), unless the next activity is home, in which case it resets to zero ($q_{i+1}=0$).
    \item Reward Function $R(s,a|m)$: For an agent $m$, the reward function $R(s,a|m)$ defines the expected reward of choosing an action $a$ given the current state $s$. To account for individual differences in behavioral preferences, we include socio-demographic features such as income, age, gender, etc., as part of the input to the reward function.
    \item Policy function $\pi(a|s,m)$:  The policy function $\pi(a|s,m)$ describes the probability distribution over actions $a$ that an agent $m$ might take given the current state $s$. While the reward function assigns a value to each state-action pair indicating human preferences, the policy function captures the behavioral patterns of humans.
\end{itemize}

\subsubsection{Adversarial IRL for activity-travel behavior modeling} \label{model:AIRL}

IRL is a learning paradigm that aims to infer a reward function to explain human preferences and subsequently learn a policy function to replicate human choice behavioral patterns. Like DDCMs, it operates under the assumption that people make decisions to maximize some reward functions, though they are unknown. However, the original IRL is ambiguous and ill-defined, as many optimal policies can explain a set of demonstrations, and many rewards can explain an optimal policy \citep{song2024state}. To address the first ambiguity, \cite{ziebart2008maximum} developed a probabilistic approach based on the principle of maximum entropy. Building on this, \cite{fu_learning_2018} further addressed the second ambiguity with adversarial IRL (AIRL), which introduced a specialized adversarial network to recover disentangled rewards that are invariant to changing dynamics. The original AIRL was proposed for robotic control tasks, and here we adapted it for modeling sequential activity-travel behavior.

\begin{figure}[!ht]
  \centering
  \includegraphics[width=\textwidth]{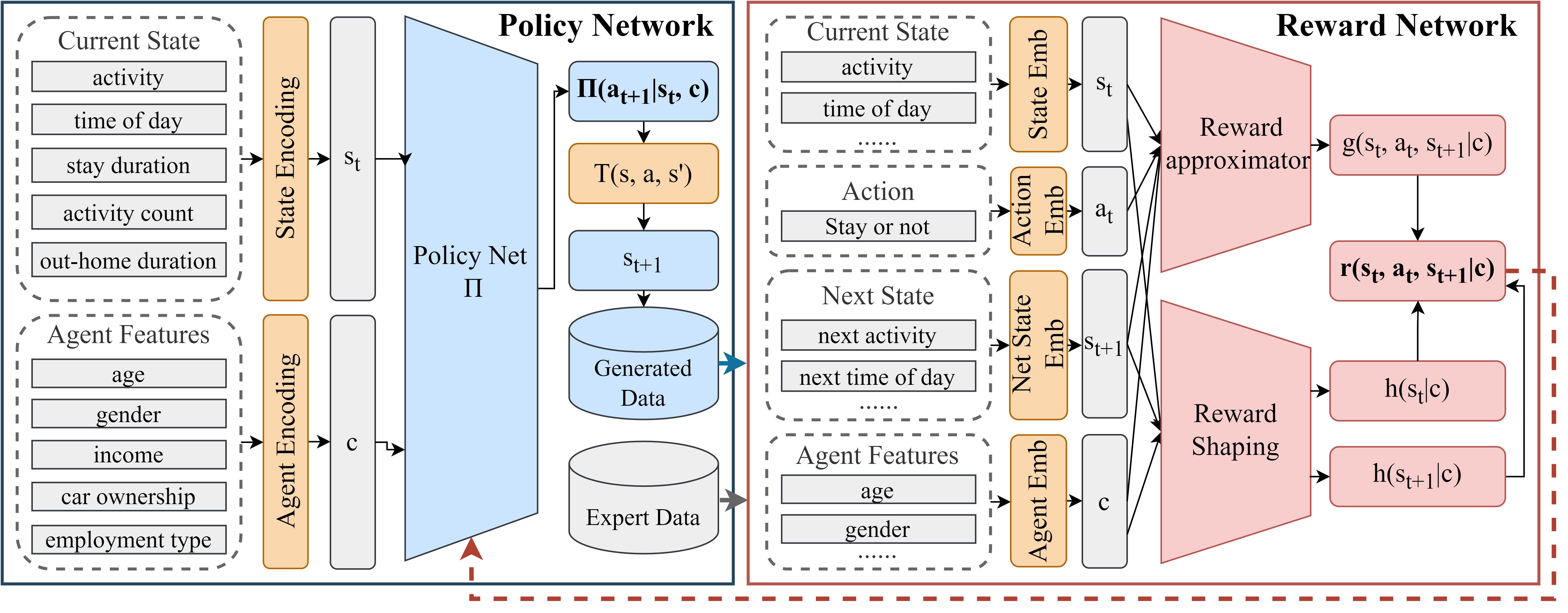}
  \caption{The network structure of AIRL for activity-travel choice modeling}\label{fig:AIRL}
\end{figure}

AIRL operates within an adversarial framework, comprising a generator and a discriminator. The generator creates realistic trajectories and learns a policy function that maps each state to a probability distribution over actions. The discriminator distinguishes between generated and observed behavior by inferring a reward function, which assigns a value representing the preference of each state-action pair. An overview of the adapted AIRL framework in this study is shown in Figure~\ref{fig:AIRL}.

%To capture complex human behavioral patterns and preferences, we use deep neural networks (DNNs) to approximate the policy and reward functions within the AIRL framework. Specifically, 
The policy network incorporates two types of input features: those related to the current state and those related to the socio-demographic characteristics of an agent. For features of the current state, which are all categorical variables, we learn an embedding matrix for each feature within the model. The matrix maps each category to a latent vector, with the matrix's width corresponding to the number of possible categories and its length to the dimension of the latent vector. For continuous positive socio-demographic features like income and age, we normalize them to a 0-1 scale before inputting them into the model. For categorical socio-demographic features such as employment type, we use embedding techniques to encode them into a latent space. All the encoded features are then concatenated and fed into a three-layer feedforward network, followed by a Softmax function to output the choice probability of the next activity. As mentioned in Section~\ref{model:MDP}, we add an action mask before the Softmax function to ensure that only feasible state-action transitions are considered.

%We also employ a DNN to approximate the reward function within the discriminator. 
The reward network incorporates the current state and individual socio-demographic features as inputs, encoded using the same methods as in the generator. Additionally, to capture the immediate reward of transitioning to the next state, the reward network takes the next state features as inputs, which are encoded similarly to the current state features. To update an unshaped reward invariant to the dynamic environment, the reward function is formulated as follows:
\begin{equation}
    R_{\theta, \phi}(s_t,a_t,s_{t+1}|m) = g_{\theta}(s_t,a_t|m) + \gamma h_{\phi}(s_{t+1}|m) - h_{\phi}(s_t|m),
\label{eq:r-reward}
\end{equation}
where $g_{\theta}(s_t,a_t|m)$ represents the immediate reward approximator. It captures the intrinsic reward the agent $m$ receives for taking action $a_t$ in state $s_t$. $h_{\phi}(s_t|m)$ and $h_{\phi}(s_{t+1}|m)$ represent the reward shaping components designed to shape the reward function to make it invariant to changing dynamics in the environment. The term $\gamma h_{\phi}(s_{t+1}|m)$ adjusts the reward based on the expected future state $s_{t+1}$, while $h_{\phi}(s_{t}|m)$ adjusts it based on the current state $s_t$. This difference helps smooth the reward landscape, making the learned reward function more robust. We use two separate DNNs to approximate $g_{\theta}(*)$ and $h_{\phi}(*)$ separately, each of which is a three-layer feed-forward network based on the encoded state and socio-demographic features. 
%At optimality, $f^*(s_t,a_t,s_{t+1}|m)$ is the advantage function of the optimal policy. 

Based on the reward and policy functions, the discriminator in AIRL is defined as follows:
\begin{equation}
    D_{\theta, \phi}(s_t,a_t,s_{t+1}|m)=\frac{\exp \big(R_{\theta, \phi}(s_t,a_t,s_{t+1}|m)\big)}{\exp \big(R_{\theta, \phi}(s_t,a_t,s_{t+1}|m)\big) + \pi(a_{t} \mid s_t,m)}.
\end{equation}
The discriminator is trained to differentiate between generated and real behavior by minimizing the cross-entropy loss between state-action pairs produced by the generator and those from observed trajectories. The policy function aims to identify an optimal policy that maximizes the expected reward as estimated by the discriminator. To achieve this, we employ a state-of-the-art policy gradient algorithm called Proximal Policy Optimization (PPO) \citep{schulman2017proximal} to train the policy network.

% The objective of the discriminator $D$ is to minimize the cross-entropy loss between generated data and actual ones:
% \begin{equation}
% \label{eq:f}
%     \min_{\theta, \phi} - E_D \left[ \log \big(D_{\theta, \phi}(s_t,a_t,s_{t+1}|c) \big)\right] - E_{\pi_G} \left[ \log \big(1 - D_{\theta, \phi}(s_t,a_t,s_{t+1}|c) \big)\right]
% \end{equation}
% where $E_D$ and $E_{\pi_G}$ indicate expected values over actual and generated trajectories, respectively.

\subsection{DIRL Interpretation} \label{method: interpretation}

The trained AIRL model infers people's activity-travel behavior patterns through the policy function and their activity-travel preferences through the reward function. By using DNNs to approximate these functions, the model can capture more complex patterns and preferences compared to traditional utility-based econometric models with linear restrictions. However, due to the black-box nature of DNNs, it is challenging to directly explain why and how these networks produce specific outcomes, leaving the ``knowledge" embedded in the policy and reward functions unclear. To address this issue, we introduce an interpretation framework to extract human-understandable knowledge from DIRL. This framework consists of two components: one for extracting behavioral patterns from the policy network, and the other for extracting preference insights from the reward network. The details of these two components are elaborated below.
% using a knowledge distillation technique \citep{hinton2015distilling}
%  using an example-based explanation technique \citep{poche2023natural}

\subsubsection{Policy interpretation}\label{method:policy_interpret}

The original policy function is represented using DNNs, making it challenging to identify the learned knowledge from individual parameters. To address this issue, we adapt a knowledge distillation method for policy interpretation, initially introduced by \cite{hinton2015distilling}. A key insight from their work is that for complex models that learn to discriminate between classes, the knowledge learned is reflected in the choice probabilities assigned to each class. %Some probabilities are significantly larger than others, and these relative probabilities reveal how the complex model generalizes. 
To transfer the knowledge from large, cumbersome models to smaller models, \cite{hinton2015distilling} proposed training a smaller model using the class probabilities produced by the large model as ``soft targets." %This method not only facilitates the wide deployment of smaller models in scenarios where large models are too computationally expensive but also improves the performance of the smaller models by distilling knowledge from the larger ones. 
We adapt this idea for interpretation purposes, using a surrogate interpretable model to distill knowledge from the policy network. Specifically, we use a multinomial logit (MNL) model as the surrogate, a common choice for choice analysis that provides comprehensive econometric interpretations \citep{wang2020deep, sifringer2020enhancing}. To extract knowledge from the policy network, the surrogate MNL model is trained to maximize the log probability of both ground-truth labels from observed behavior and ``soft labels" from the policy network:
\begin{equation}
L_{\theta} = -\sum_{m=1}^M{\sum_{t=1}^{N}{(1 - \alpha) \times \bm{y}_{t,m} \log(\hat{\bm{y}}^{mnl}_{t,m}) + \alpha \times \hat{\bm{y}}^{irl}_{t,m} \log(\hat{\bm{y}}^{mnl}_{t,m})}},
\end{equation}
where $M$ is the number of users in the training set, $N$ is the number of timesteps. $\bm{y}_{t,m}$ is a one-hot vector denoting the ground-truth activity of user $m$ at time $t$, $\hat{\bm{y}}^{irl}_{t,m}$ is the choice probability vector generated by the pre-trained policy network, and $\hat{\bm{y}}^{mnl}_{t,m}$ is the estimated choice probability of the MNL model. $\alpha \in [0, 1]$ is a trade-off weight to balance different losses. When $\alpha = 0$, the MNL model is trained completely based on ground-truth labels, while when $\alpha = 1$, the model is trained completely based on soft labels generated by the policy network. The MNL model can then be directly interpreted through its learned parameters. 

\subsubsection{Reward interpretation}\label{method:reward_interpret}

The goal of DIRL is to infer an unknown reward function that explains how people make decisions based on a set of demonstration data. A natural approach for interpreting rewards is to map trajectories in the demonstration data to their corresponding reward values using the pre-trained reward network. %This approach falls under example-based explanation methods in explainable artificial intelligence (XAI) \citep{poche2023natural}, which interpret machine learning models by providing specific examples from the demonstration data. 
By doing so, we establish a direct link between human behavioral patterns and underlying preferences. The reward function assigns a value to each state-action pair, indicating the immediate utility an agent can gain. By applying the pre-trained reward network to each timestep, we can map an activity-travel sequence $Y_m$ to a reward sequence $R_m = \{r_{1,m}, r_{2,m},...r_{N,m}\}$ where $r_{i,m}$ is the inferred reward at the $i$-th timestep for agent $m$. The reward sequence provides a dynamic view of underlying human preferences in sequential behavioral processes. It also enables us to convert discrete behavioral sequences into continuous values, facilitating further analysis.

People can have varying preferences when making sequential activity-travel decisions. For instance, some may have regular, repetitive behavioral patterns, while others explore new and diverse activities with less predictable patterns \citep{pappalardo2015returners}. To identify different types of decision-makers, we use a clustering technique to group individuals based on their corresponding reward sequences. People within the same cluster share similar preferences for activity-travel planning. In this case, without loss of generalizability, we employ the k-means algorithm \citep{krishna1999genetic}, a popular method for partitioning data into $k$ distinct clusters. %K-means works by iteratively assigning each data point to the nearest cluster centroid and then updating the centroids based on the mean of the points in each cluster. This process continues until convergence, resulting in clusters that capture the inherent similarities in the data (reward patterns in our case).

%Note that this is in some sense inspired from SHAP clustering \cite{clement2024beyond}, which clusters data samples based on their associated SHAP values from different features so that similar samples with similar distribution of feature importance can be identified.

The reward values estimated from the reward function represent the immediate utility an individual can gain. In sequential decision-making, it is important to consider not only the immediate reward but also the long-term expected utility. %However, not all individuals can perfectly maximize this long-term expected utility. 
The reward function allows us to quantify the long-term expected return $U^{(i)}_m$ for an agent $m$ starting from timestep $i$ as follows:
\begin{equation}
    U^{(i)}_m = \sum_{t=i}^{N}{\gamma^{t-i} r_{t,m}}.
\label{eq:f-reward}
\end{equation}
In our case, without loss of generality, $i$ is set to 1, representing the accumulated return starting from midnight. This long-term expected utility can compress the reward sequence into a single value, representing the overall utility an individual gains from the entire sequential planning process. Using this quantified long-term utility, we can compare the ``values" of different activity-travel patterns. This comparison enables us to understand which decision-making process is optimal, at least from the perspective of the DIRL model. %This insight can potentially enhance individual scheduling efficiency and lead to better social outcomes.

\section{Experiments}
% \blindtext

This section demonstrates the feasibility of extracting behavioral patterns and preference knowledge from a well-trained DIRL model. We start by introducing the dataset used for analysis (see Section~\ref{exp:data}) and then compare the prediction performance of different methods, showcasing the effectiveness of DIRL in accurately mimicking observed activity-travel patterns (see Section~\ref{exp:accuracy}). Next, we present the extracted behavioral knowledge based on the policy network in Section~\ref{exp:policy_interpret} and the preference knowledge derived from the reward network in Section~\ref{exp:reward_interpret}.

\subsection{Dataset}\label{exp:data}

\begin{figure}[!ht]
  \centering
  \includegraphics[width=\textwidth]{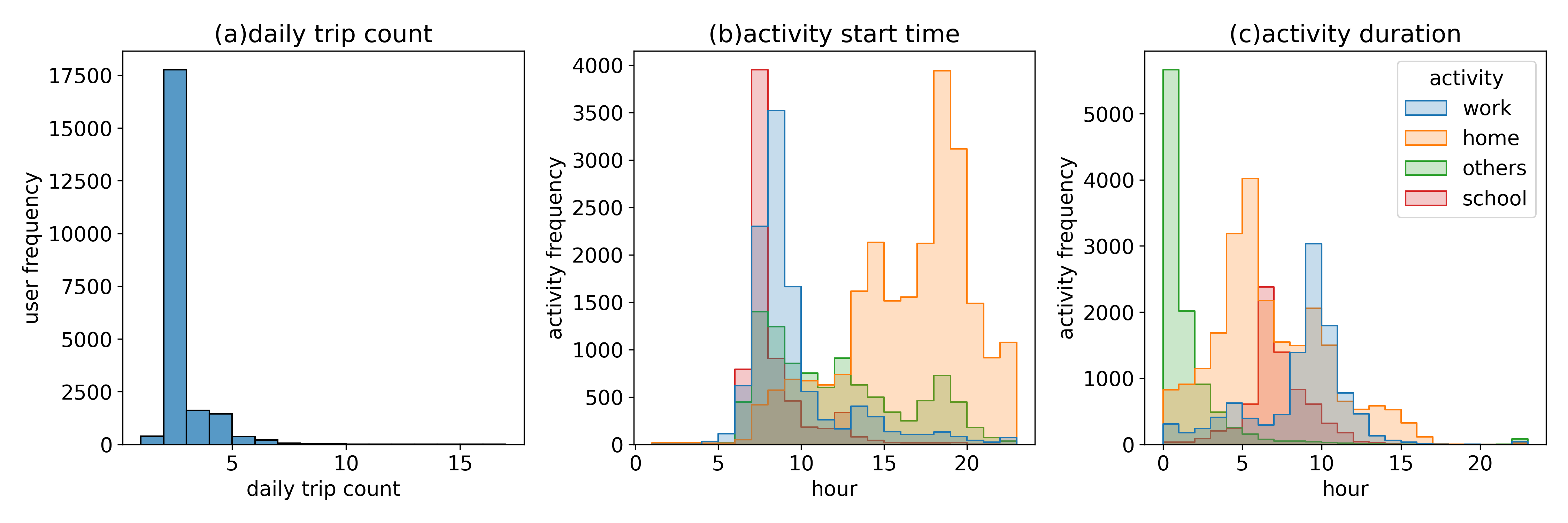}
  \caption{Singapore travel survey data}\label{fig:data}
\end{figure}

We use the Household Travel Survey data from Singapore as a case study. This dataset, provided by the Singapore Land Transport Authority, contains detailed trip information for 35,714 users over a self-reported day between June 25, 2012, and May 30, 2013. Each user reports the purpose of their trips along with start and end times, allowing us to construct complete daily activity-travel sequences for each individual. We categorize activities into four types: home, work, school, and others. The survey also includes socio-demographic features such as age, income, car ownership, gender, and employment type. After filtering out users with incomplete or invalid trip or socio-demographic information, we are left with 50,807 trips from 21,936 users. The stay duration for most other activities is quite short, typically within 1 hour. To formulate the activity-travel decision process into a MDP, we divide a day into 96 15-minute intervals. This results in each individual having 96 decision steps per day. Figure~\ref{fig:data} provides an intuitive illustration of the data, showing the distribution of the number of trips per day, as well as the start times and duration of various activities.
%Figure~\ref{fig:data}(a) illustrates the distribution of the number of trips per day among different individuals. Most users have 2 trips per day, while a small number have high trip frequencies, up to 16 trips in a day. Figure~\ref{fig:data}(b) shows the start times for different activity types. There is a morning peak for trips to work and school, while trips returning home are more spread out after 2 PM, with an evening peak at 6 PM. The start times for other activities are more randomly distributed throughout the day. Figure~\ref{fig:data}(c) shows the stay duration for different activities. Most people spend around 9 hours at work and 6 hours at school. 

\subsection{Prediction accuracy of different models}\label{exp:accuracy}

To demonstrate the effectiveness of DIRL in modeling sequential activity-travel behavior, we compare its prediction performance with several baseline methods:
\begin{itemize}
    \item Mobility Markov Chain (MC) \citep{gambs2012next}: MMC treats sequential behavior as a discrete stochastic process, assuming that the probability of moving to the next state depends only on the current state. For activity-travel behavior modeling, each state is defined by the current time step and current activity. The model learns the probability of the next activity based on the current state from the training data. Due to the inherent model restrictions of MMC, it is challenging to incorporate socio-demographic contexts as input.
    \item Multinomial Logit Model (MNL): MNL assumes that individuals choose behaviors to maximize utility values and estimates a utility function to model choice probabilities. In this context, each state-action pair is treated as an independent data sample, and MNL is estimated based on the activity labels at the next time step. The utility function in MNL takes current state features and individual socio-demographic context as input, similar to the policy function in DIRL.
    \item Behavior Cloning (BC): BC models sequential behavior as a supervised learning problem. It trains the model to map states to actions using ground-truth activity labels from expert demonstrations. For a fair comparison, BC adopts the same structure as the policy network in the DIRL model. The key difference is that the policy network of BC is directly trained to maximize the log-probability of the action labels, whereas that of DIRL is trained to represent the optimal policy that maximizes the learned reward function.
\end{itemize}

We randomly sampled 80\% of users as the training set and the remaining 20\% as the test set. This resulted in 17,549 users (or trajectories) for model training and 4,387 users (or trajectories) for model testing. During model testing, each user starts from a virtual \textit{start} state and recursively generates predicted activities at each time step by sampling from the learned choice probabilities until the end of the day. To quantify model performance in terms of recovering real individual activity-travel patterns, we use the following metrics:
\begin{itemize}
    \item Accuracy: This metric evaluates the ratio of correctly predicted activity labels to the total number of timesteps in the dataset:
    \begin{equation}
        Accuracy = \frac{1}{M}\sum_{m=1}^{M}\sum_{i=1}^{N}\frac{1(y_{i,m}, \hat{y}_{i,m})}{N},
    \end{equation}
    where $M$ is the number of users in the test set, $N$ is the number of timesteps in a day, $y_{m,t}, \hat{y}_{m,t}$ are the ground-truth and predicted activity labels for user $m$ at timestep $i$, and $1(*)$ is a binary function that equals 1 when $y_{m,t} =\hat{y}_{m,t}$, and 0 otherwise.
    % \item F1 Score: F1 score considers both precision and recall, providing a balance between the two. It is calculated as:
    % \begin{equation}
    %     F1 Score = 2 \times \frac{1}{M}\sum_{m=1}^{M} \frac{Precision_m \times Recall_m}{Precision_m + Recall_m},
    % \end{equation}
    % where $Precision_m$ is the ratio of correctly predicted positive 
    \item Edit Distance (ED). Edit distance quantifies how dissimilar two sequences are and is computed as: 
    \begin{equation}
    ED = \frac{1}{M}\sum_{m=1}^{M}{\min \left(\frac{Edit(\hat{Y}_m, Y_m)}{N}, 1\right)},
    \end{equation}
    where $\hat{Y}_m, Y_m$ are the predicted and ground-truth activity-travel sequence for user $m$, 
    and $Edit(*)$ denotes a function counting the minimum number of operations required to transform one sequence into the other.
    \item BiLingual Evaluation Understudy score (BLEU). Originally used in machine translation, the BLEU score measures the similarity of a candidate text to reference texts. In our context, it is used to measure trajectory similarities. Higher BLEU scores indicate closer matches to the reference trajectories. BLEU compares $n$-gram matches between each predicted and ground-truth activity-travel sequence in the test set:
    \begin{equation}
        {BLEU}_n = \frac{1}{M}\sum_{m=1}^{M}{(\prod_{j=1}^{n}P_j)^{\frac{1}{n}}},
    \end{equation}
    where $P_j$ is the precision of $j$-gram matches, defined as $P_j = \frac{\sum_{c \in C_j} \min(w_c, w_{c, max})}{W}$, $C_j$ is the set of unique $j$-gram chunks found in the predicted sequence, $w_c$ is the number of occurrences of chunk $c$ in the predicted sequence; $w_{c,max}$ is the maximum number of occurrences of chunk $c$ in the ground-truth sequence, and $W$ is the total number of chunks in the predicted sequence.
\end{itemize}

Table~\ref{table:ind_perform} presents a performance comparison between DIRL and the baseline models, clearly showing that DIRL significantly outperforms all baselines in modeling activity-travel behavioral patterns. Among the baselines, the Markov Chain performs relatively poorly, which is expected since it relies solely on state-action transition probabilities from the training data to simulate possible trajectories in the test data, neglecting individual differences arising from socio-demographic features. MNL performs better by capturing context-specific factors and individual variations through the estimation of a utility function. However, its linear function assumption limits its ability to capture more complex behavioral patterns. BC outperforms MNL, likely because it utilizes deep learning architectures to capture more intricate relationships between input contextual features and choice probabilities. However, BC treats each state-action pair as an independent data sample, which restricts its capacity to capture the sequential linkage between time steps in decision-making. In contrast, AIRL excels at capturing long-term rewards by assuming that individuals are forward-looking when making sequential decisions. This makes AIRL a highly effective framework for modeling individual activity-travel behavior, enabling further interpretation of the behavioral patterns and preferences learned from the model. 

\begin{figure}[!ht]
  \centering
  \includegraphics[width=\textwidth]{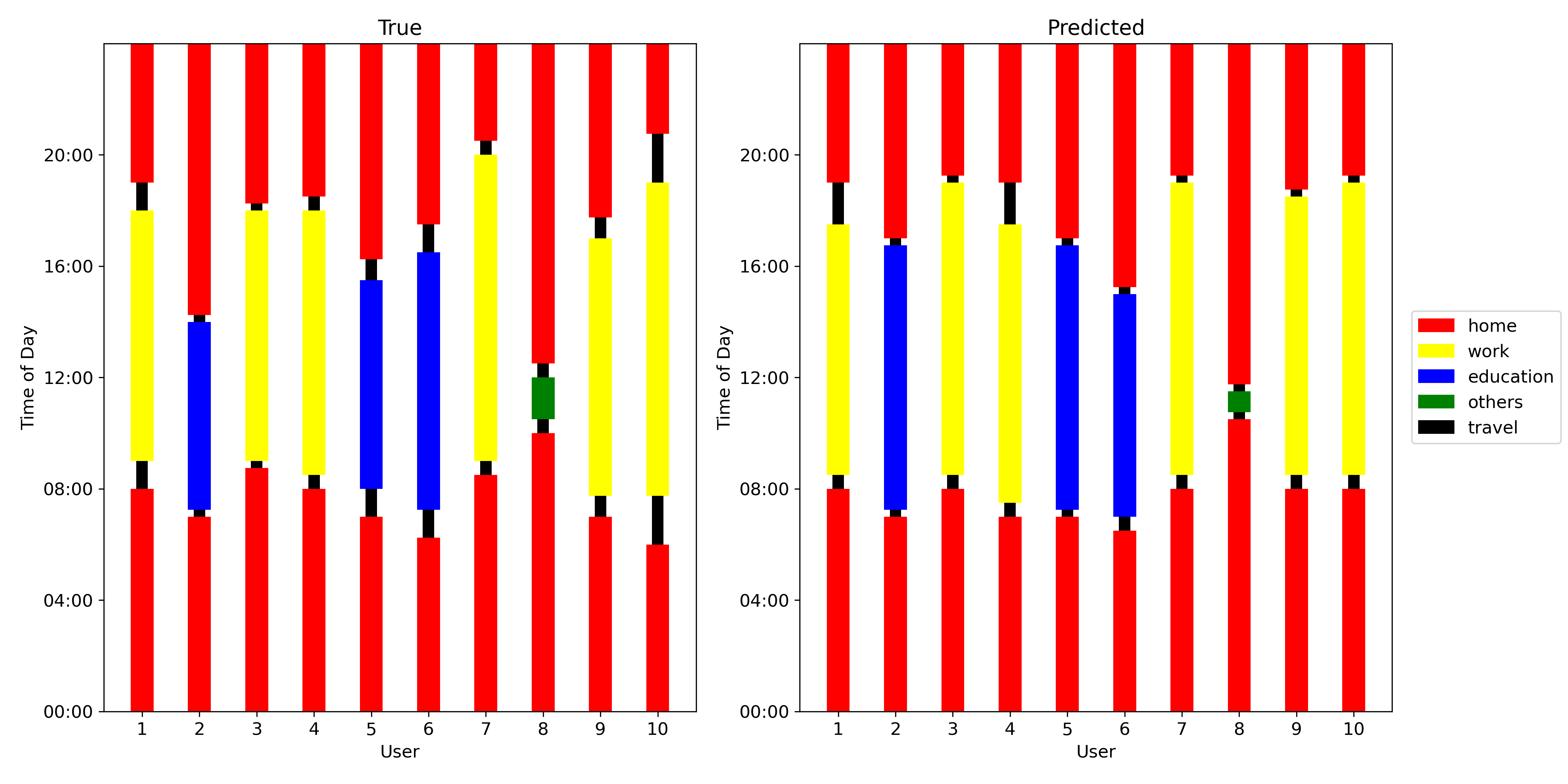}
  \caption{Examples of ground-truth and generated activity-travel sequences}\label{fig:true_predict_examples}
\end{figure}

To intuitively illustrate the effectiveness of DIRL in mimicking real-world activity-travel behaviors, Figure~\ref{fig:true_predict_examples} presents several examples of ground-truth and generated activity-travel sequences. These examples demonstrate AIRL's ability to largely replicate activity type choice and scheduling patterns. Although there are some inconsistencies regarding activity transition times, these discrepancies are acceptable given the challenging nature of the task. This complexity arises from generating the entire activity sequence recursively by the model when only individual socio-demographic features are known.

\begin{table}[ht!]
  \centering \footnotesize
  \caption{Comparison of prediction performance using different models}
    % \resizebox{\linewidth}{!}{%
    \begin{tabular}{cccc}
    %\addlinespace
    \toprule
    \multirow{2}{*}{Models} & \multicolumn{3}{c}{Metrics}\\
    & $ACC$ & $ED$ & $BLEU$\\
    \midrule
    %Markov Chain & 0.528 & 0.209 & 0.468 & 0.617\\
    MMC & 0.603 & 0.394 & 0.652\\
    MNL & 0.749 & 0.243 & 0.838\\
    BC & 0.773 & 0.217 & 0.863\\
    DIRL & \textit{\underline{0.804}} & \textit{\underline{0.189}} & \textit{\underline{0.876}}\\
    \bottomrule
    \end{tabular}%
    % }
  \label{table:ind_perform}%
\end{table}%

\subsection{Policy interpretation}\label{exp:policy_interpret}

As described in Section~\ref{method:policy_interpret}, we utilize a knowledge distillation method to uncover the behavioral patterns learned by the policy network. Specifically, we train a surrogate interpretable Multinomial Logit (MNL) model using the soft labels generated by the policy network. In this section, we first demonstrate the effectiveness of this knowledge distillation method by discussing the impact of incorporating these soft labels on the modeling performance of the MNL (see Section~\ref{res:soft_label_accuracy}). We then analyze the uncovered behavioral patterns by examining the coefficients learned by the MNL (see Section~\ref{res:mnl_interpret}). 

\subsubsection{Prediction accuracy of MNL with soft labels from the policy network}\label{res:soft_label_accuracy}

Table~\ref{table:performance_soft_labels} presents the performance of the MNL model trained with and without soft labels generated by the policy network. The results indicate that incorporating soft labels in the loss function enhances the MNL's performance across all evaluation metrics. This is because compared to using ground-truth activities as labels, the soft labels generated by the policy network provide richer information about the relative choice probabilities among different activities. Surprisingly, when $\alpha$ is set to 1, meaning the MNL is trained entirely based on the soft labels predicted by the policy network, it achieves the highest performance in terms of accuracy and edit distance, and similarly strong performance in BLEU score compared to other $\alpha$ settings. This suggests that even without the ground-truth activities, the choice probabilities learned by the policy network can effectively capture real activity-travel behaviors.

\begin{table}[ht!]
  \centering \footnotesize
  \caption{Performance comparison of MNL trained with and without soft labels from the policy network}
    % \resizebox{\linewidth}{!}{%
    \begin{tabular}{ccccc}
    %\addlinespace
    \toprule
    \multicolumn{2}{c}{\multirow{2}{*}{Models}} & \multicolumn{3}{c}{Metrics}\\
    & & $ACC$ & $ED$ & $BLEU$\\
    \hline
    Without soft labels & $\alpha$=0 & 0.750 & 0.241 & 0.837\\
    \hline
    \multirow{5}{*}{With soft labels} & $\alpha$=0.5 & 0.760 & 0.231 & 0.844\\
    & $\alpha$=0.9 & 0.771 & 0.220 & \underline{\textit{0.846}}\\
    & $\alpha$=0.99 & 0.772 & 0.220 & 0.842\\
    & $\alpha$=1 & \underline{\textit{0.775}} & \underline{\textit{0.217}} & 0.844\\
    \bottomrule
    \end{tabular}%
    % }
  \label{table:performance_soft_labels}%
\end{table}%

% \begin{table}[ht!]
%   \centering \footnotesize
%   \caption{Performance comparison of MNL trained with and without soft labels from the policy network}
%     % \resizebox{\linewidth}{!}{%
%     \begin{tabular}{cccccc}
%     %\addlinespace
%     \toprule
%     \multicolumn{2}{c}{\multirow{2}{*}{Models}} & \multicolumn{4}{c}{Metrics}\\
%     & & $ACC$ & $F1$ & $ED$ & $BLEU$\\
%     \hline
%     Without soft labels & $\alpha$=0 & 0.750 & 0.498 & 0.241 & 0.837\\
%     \hline
%     \multirow{5}{*}{With soft labels} & $\alpha$=0.5 & 0.760 & 0.505 & 0.231 & 0.844\\
%     & $\alpha$=0.9 & 0.771 & \underline{\textit{0.509}} & \underline{\textit{0.220}} & \underline{\textit{0.846}}\\
%     & $\alpha$=0.99 & 0.772 & 0.504 & \underline{\textit{0.220}} & 0.842\\
%     & $\alpha$=1 & \underline{\textit{0.775}} & 0.507 & 0.217 & 0.844\\
%     \bottomrule
%     \end{tabular}%
%     % }
%   \label{table:performance_soft_labels}%
% \end{table}%

% \begin{figure}[t!]
% \captionsetup[subfigure]{justification=centering}
% \centering
% \subfloat[Zero-one loss (20 Var)]{\includegraphics[width=0.25\linewidth]{{experiment/sce3_run1_accuracy}.png}\label{sfig:s3_acc_20_var}}
% \subfloat[Log loss (20 Var)]{\includegraphics[width=0.25\linewidth]{{experiment/sce3_run1_logloss}.png}\label{sfig:s3_log_20_var}}
% \subfloat[Function approximation loss (20 Var)]{\includegraphics[width=0.25\linewidth]{{experiment/sce3_run1_func_approx}.png}\label{sfig:s3_est_20_var}} 
% \caption{Scenario 3. Comparison of DNN, BNL, and BXL for zero-one, log, and function approximation losses.}
% \label{fig:sce3}
% \end{figure}

\subsubsection{Interpretation of the surrogate MNL model}\label{res:mnl_interpret}

This section analyzes the behavioral patterns derived from the policy network by examining the coefficients of the surrogate MNL model. To ensure the robustness of our interpretations, we employ a bootstrapping strategy. Specifically, we randomly sample 80\% of the observations with replacement to train the surrogate MNL model, repeating this process 100 times. Table 3 presents the average coefficients and the [0.025, 0.975] confidence intervals for each input variable based on the 100 bootstrap samples. Variables with inconsistent directional impacts within these intervals are highlighted in grey. The results show that most variables maintain consistent directional impacts over 95\% of the time, validating the robustness of our interpretations. Features with consistent directional impacts in most experiments are likely to significantly influence human activity and travel behavior patterns, while those with inconsistent impacts may be less significant in explaining these patterns.

%\newcolumntype{Y}{>{\raggedright\arraybackslash}X}
\afterpage{
    \clearpage
    \newgeometry{top=1in, bottom=1in, left=0.5in, right=0.5in}
    \newcolumntype{Y}{>{\raggedright\arraybackslash}X}
    % Landscape page with a wide table
    \begin{landscape}
        \begin{table}[h]
            \centering
            \scriptsize
            \setlength{\extrarowheight}{2pt} % Increase row height
            \begin{tabularx}{\linewidth}{|l|YYY|YYY|YYY|YYY|YYY|}
                \hline
                \multirow{2}{*}{Variables} & \multicolumn{3}{c|}{Home} & \multicolumn{3}{c|}{Work} & \multicolumn{3}{c|}{School} & \multicolumn{3}{c|}{Others}  & \multicolumn{3}{c|}{Travel} \\
                \cline{2-16}
                & Coef & [0.025 & 0.975] & Coef & [0.025 & 0.975] & Coef & [0.025 & 0.975] & Coef & [0.025 & 0.975]  & Coef & [0.025 & 0.975]\\
                \hline
    Activity: Home	&	3.518	&	3.425	&	3.608	&	-16.171	&	-18.927	&	-12.800	&	-15.818	&	-18.470	&	-12.436	&	-14.547	&	-16.955	&	-11.652	&	-0.689	&	-0.749	&	-0.642	\\
    Activity: Work	&	-16.316	&	-18.872	&	-13.219	&	3.765	&	3.635	&	3.866	&	-9.523	&	-12.497	&	-6.576	&	-13.766	&	-15.625	&	-11.495	&	-0.900	&	-0.993	&	-0.829	\\
    Activity: School	&	-15.835	&	-17.993	&	-12.960	&	-11.909	&	-13.712	&	-9.671	&	3.589	&	3.503	&	3.656	&	-13.596	&	-15.235	&	-11.606	&	-0.632	&	-0.693	&	-0.577	\\
    Activity: Others	&	-16.026	&	-18.636	&	-13.460	&	-13.304	&	-15.123	&	-11.115	&	-11.936	&	-13.612	&	-9.493	&	3.532	&	3.419	&	3.623	&	0.847	&	0.807	&	0.891	\\
    Activity: Travel	&	-2.101	&	-2.205	&	-1.977	&	-0.687	&	-0.794	&	-0.589	&	-1.054	&	-1.126	&	-0.976	&	0.597	&	0.493	&	0.752	&	1.415	&	1.359	&	1.492	\\
    \hline
    Time: 0-2h	&	4.258	&	3.492	&	4.723	&	-0.944	&	-1.397	&	-0.716	&	-1.409	&	-1.429	&	-1.379	&	-0.957	&	-1.338	&	-0.720	&	-3.984	&	-4.614	&	-3.784	\\
    Time: 2-4h	&	3.035	&	2.804	&	3.154	&	-1.123	&	-1.361	&	-0.545	&	-1.373	&	-1.400	&	-1.353	&	-0.259	&	\textcolor{gray}{-0.975}	&	\textcolor{gray}{0.297}	&	-4.008	&	-4.172	&	-3.907	\\
    Time: 4-6h	&	0.512	&	0.388	&	0.567	&	0.150	&	0.004	&	0.308	&	-0.309	&	\textcolor{gray}{-0.621}	&	\textcolor{gray}{0.042}	&	-0.822	&	-1.389	&	-0.370	&	-1.100	&	-1.343	&	-0.973	\\
    Time: 6-8h	&	-1.414	&	-1.476	&	-1.369	&	0.738	&	0.612	&	0.841	&	1.861	&	1.795	&	1.911	&	-0.560	&	-0.698	&	-0.422	&	0.257	&	0.209	&	0.292	\\
    Time: 8-10h	&	-1.307	&	-1.437	&	-1.216	&	1.029	&	0.888	&	1.143	&	1.455	&	1.380	&	1.582	&	-0.608	&	-0.675	&	-0.533	&	-0.214	&	-0.248	&	-0.179	\\
    Time: 10-12h	&	-0.718	&	-0.836	&	-0.630	&	0.613	&	0.487	&	0.725	&	0.657	&	0.583	&	0.759	&	-0.487	&	-0.542	&	-0.364	&	-0.455	&	-0.489	&	-0.397	\\
    Time: 12-14h	&	0.183	&	0.137	&	0.262	&	0.512	&	0.458	&	0.612	&	-0.493	&	-0.609	&	-0.413	&	-0.249	&	-0.313	&	-0.177	&	-0.205	&	-0.232	&	-0.143	\\
    Time: 14-16h	&	0.482	&	0.428	&	0.550	&	0.213	&	0.139	&	0.338	&	-0.892	&	-0.984	&	-0.755	&	-0.361	&	-0.445	&	-0.291	&	-0.069	&	-0.112	&	-0.019	\\
    Time: 16-18h	&	0.603	&	0.517	&	0.648	&	-0.331	&	-0.427	&	-0.188	&	-0.912	&	-0.990	&	-0.861	&	-0.082	&	\textcolor{gray}{-0.135}	&	\textcolor{gray}{0.007}	&	0.280	&	0.234	&	0.329	\\
    Time: 18-20h	&	0.698	&	0.639	&	0.774	&	-1.054	&	-1.161	&	-0.851	&	-1.263	&	-1.374	&	-1.064	&	0.115	&	0.049	&	0.179	&	0.202	&	0.163	&	0.258	\\
    Time: 20-22h	&	0.672	&	0.558	&	0.711	&	-0.943	&	-1.109	&	-0.684	&	-1.134	&	-1.379	&	-0.831	&	-0.382	&	-0.466	&	-0.262	&	-0.031	&	\textcolor{gray}{-0.079}	&	\textcolor{gray}{0.020}	\\
    Time: 22-24h	&	0.337	&	0.249	&	0.422	&	-0.413	&	-0.638	&	-0.207	&	-0.171	&	\textcolor{gray}{-1.121}	&	\textcolor{gray}{0.370}	&	-0.731	&	-0.937	&	-0.606	&	-0.015	&	\textcolor{gray}{-0.106}	&	\textcolor{gray}{0.090}	\\
    \hline
    Stay Duration	&	-3.493	&	-3.783	&	-3.048	&	1.145	&	0.896	&	1.515	&	-0.528	&	-1.011	&	-0.006	&	3.669	&	3.145	&	4.196	&	2.191	&	2.023	&	2.334	\\
    Activity Count	&	13.318	&	10.545	&	17.252	&	-14.071	&	-18.431	&	-11.173	&	-7.338	&	-8.765	&	-5.779	&	-4.243	&	-5.841	&	-2.602	&	-1.982	&	-2.339	&	-1.491	\\
    Out-home Duration	&	3.625	&	3.203	&	4.134	&	-2.932	&	-3.450	&	-2.493	&	-1.732	&	-2.133	&	-1.398	&	-0.581	&	-0.882	&	-0.240	&	1.290	&	1.191	&	1.452	\\
    \hline
    Age	&	-0.055	&	\textcolor{gray}{-0.135}	&	\textcolor{gray}{0.001}	&	0.160	&	0.075	&	0.253	&	0.387	&	\textcolor{gray}{-0.064}	&	\textcolor{gray}{1.020}	&	0.191	&	0.067	&	0.327	&	0.108	&	\textcolor{gray}{-0.024}	&	\textcolor{gray}{0.188}	\\
    Gender (1: female)	&	0.049	&	0.007	&	0.085	&	-0.188	&	-0.244	&	-0.107	&	-0.222	&	-0.288	&	-0.161	&	0.117	&	0.036	&	0.223	&	-0.133	&	-0.196	&	-0.085	\\
    Car Ownership	&	0.032	&	\textcolor{gray}{-0.056}	&	\textcolor{gray}{0.096}	&	0.095	&	0.000	&	0.182	&	-0.013	&	\textcolor{gray}{-0.164}	&	\textcolor{gray}{0.071}	&	0.074	&	\textcolor{gray}{-0.010}	&	\textcolor{gray}{0.161}	&	0.037	&	\textcolor{gray}{-0.043}	&	\textcolor{gray}{0.090}	\\
    Income	&	0.463	&	0.333	&	0.659	&	0.224	&	0.105	&	0.416	&	-3.738	&	-4.404	&	-3.127	&	0.607	&	\textcolor{gray}{-0.074}	&	\textcolor{gray}{0.989}	&	0.092	&	0.022	&	0.206	\\
    \hline
    Homemaker	&	1.298	&	1.202	&	1.400	&	-2.354	&	-2.480	&	-2.217	&	-3.644	&	-6.051	&	-1.731	&	0.187	&	0.052	&	0.332	&	-0.317	&	-0.380	&	-0.231	\\
    Full-time Student	&	0.266	&	0.198	&	0.332	&	-1.627	&	-1.791	&	-1.500	&	1.609	&	1.533	&	1.674	&	-0.526	&	-0.605	&	-0.460	&	-0.096	&	-0.196	&	-0.042	\\
    Employed Part-time	&	0.382	&	0.268	&	0.470	&	0.860	&	0.781	&	0.937	&	-2.545	&	-2.860	&	-2.183	&	-0.662	&	-0.763	&	-0.522	&	-0.242	&	-0.337	&	-0.160	\\
    Employed Full-time	&	0.085	&	0.009	&	0.190	&	1.490	&	1.405	&	1.556	&	-3.204	&	-3.440	&	-3.026	&	-0.814	&	-0.943	&	-0.627	&	0.002	&	\textcolor{gray}{-0.051}	&	\textcolor{gray}{0.083}	\\
    Self-employed	&	0.444	&	0.346	&	0.527	&	0.915	&	0.821	&	0.995	&	-1.548	&	-2.013	&	-1.212	&	-0.880	&	-1.226	&	-0.633	&	-0.247	&	-0.384	&	-0.137	\\
    Unemployed	&	0.995	&	0.946	&	1.093	&	-0.604	&	-0.767	&	-0.498	&	-1.804	&	-4.077	&	-0.682	&	0.041	&	\textcolor{gray}{-0.055}	&	\textcolor{gray}{0.184}	&	-0.422	&	-0.565	&	-0.315	\\
    Retired	&	1.234	&	1.145	&	1.326	&	-0.932	&	-1.083	&	-0.780	&	-4.377	&	-4.890	&	-3.982	&	0.169	&	\textcolor{gray}{-0.014}	&	\textcolor{gray}{0.308}	&	-0.456	&	-0.586	&	-0.375	\\
    Domestic Worker	&	1.082	&	0.930	&	1.261	&	-1.439	&	-2.272	&	-0.829	&	-2.667	&	-2.897	&	-2.316	&	-0.133	&	\textcolor{gray}{-0.295}	&	\textcolor{gray}{0.011}	&	-0.524	&	-0.655	&	-0.430	\\
            \hline
            \end{tabularx}
            \label{table:mnl_coef}
            \caption{Coefficients of the surrogate MNL model}
        \end{table}
    \end{landscape}
    
    \restoregeometry % Restore original margins after landscape page
    \clearpage % Ensure the document continues properly
}

The current activity an agent is engaged in significantly influences their next activity choice. When the next activity is home, work, or education, there are positive coefficients for the same activity in the current time step. This means individuals tend to continue with the same activity in the next time step. Conversely, when the current activity falls under the category of others, choosing travel in the next time step is associated with positive coefficients, indicating that individuals are more likely to transition to travel. This pattern is reasonable, as the duration spent at home, work, and education is typically longer, whereas other activities generally have shorter durations, as illustrated in Figure~\ref{fig:data}.

%The time of day when an individual makes decisions also significantly impacts activity choices. 
By examining the time slots with positive coefficients, we can identify the most common periods for different activities. Home activities are most common early in the morning (0-6h) and in the afternoon and evening (14-24h). The most common work hours are between 6-16h, while school activities peak from 6-12h. For other activities, all time intervals show negative coefficients, likely because these activities occupy a smaller portion of individuals' daily lives. Travel is more frequent during 6-8h and 16-20h, corresponding to the morning and evening peak hours for commuting. %This finding also suggests that morning peak hours are more concentrated, while evening peak hours are more dispersed. This dispersion is reasonable, as the end times for work or school vary widely, and people are more likely to engage in other activities on their way home.

The duration of staying at the current activity positively correlates with choosing travel as the next activity. This is reasonable as people are more likely to move on to another activity after spending a significant amount of time in one place. The number of activities since the start of the day shows positive coefficients for home and negative coefficients for other activities. This indicates that as the number of activities increases, people are more likely to return home and less likely to engage in out-of-home activities such as work, school, and others. Additionally, the duration of out-of-home activities is positively correlated with both travel and home, while negatively correlated with work, school, and other activities. This further supports the idea that as out-of-home duration increases, people tend to travel back home.

Behavioral patterns vary across different age groups. Increased age is positively correlated with work and other activities, indicating that older individuals are more likely to engage in these activities. This is reasonable, as older people often have more social and family responsibilities. Gender differences also emerge in activity-travel planning. Females are positively correlated with staying at home and engaging in other activities, while negatively correlated with work, school, and travel. This suggests that females are more likely to stay at home or participate in other activities and less likely to go to work or school, possibly due to a greater share of household duties. Income level also influences activity choices. High-income individuals are more likely to go to work and less likely to go to school, which is expected as schoolgoers typically do not have an income.

Different employment types clearly correspond to distinct activity planning patterns. Homemakers are more likely to stay at home or engage in other activities rather than work, school, or travel. This is reasonable as they often take more household responsibilities, such as shopping and picking up children. Full-time students tend to go to school, while full-time, part-time, and self-employed individuals are more likely to go to work. Notably, the coefficient for travel among full-time employees is higher than those for part-time and self-employed individuals, indicating that full-time employees are more likely to travel compared to part-time and self-employed individuals. This difference suggests that work modes influence travel tendencies: full-time employees have fixed schedules, resulting in regular commuting trips, whereas part-time and self-employed individuals have more flexibility in their time and location choices. Retired and unemployed individuals, as well as domestic workers, show positive coefficients for staying at home and negative coefficients for work, school, or travel. This indicates that these individuals tend to stay at home, which is reasonable since they do not need to go to office or school and have fewer travel requirements.

\subsection{Reward interpretation}\label{exp:reward_interpret}

DIRL uncovers a reward function that maps state-action pairs to human-perceived preferences. %Similar to utility functions in econometric models, reward values in IRL represent the immediate benefit an individual gains by choosing a specific activity at a given time. %To interpret the reward function learned by the deep reward network, we employ an example-based analysis method, as described in Section~\ref{method:reward_interpret}. 
Using the pre-trained reward network, we can estimate a reward value for each step in an activity sequence. This allows us to analyze both short-term rewards and long-term returns of different activity-travel patterns as described in Section~\ref{method:reward_interpret}.

\subsubsection{Short-term reward}\label{res:short_reward}

\begin{figure}[!ht]
  \centering
  \includegraphics[width=\textwidth]{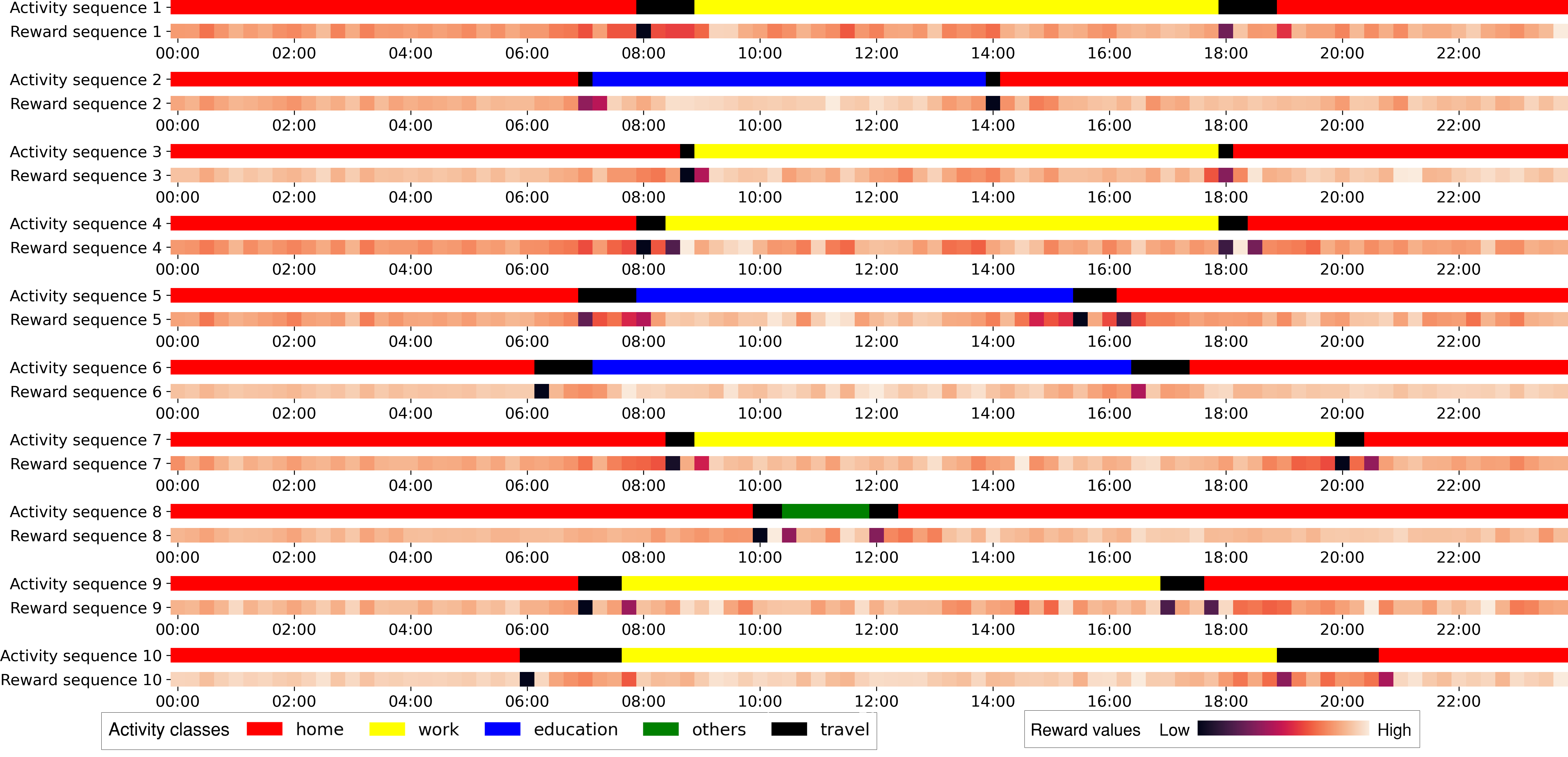}
  \caption{Examples of activity-travel sequence and corresponding reward sequence}\label{fig:reward_examples}
\end{figure}

Figure~\ref{fig:reward_examples} provides an intuitive illustration of the learned reward values at different steps in example activity-travel trajectories. Notably, the steps where individuals transition from an activity to travel are typically associated with the lowest instant rewards in the trajectory. This aligns with the intuition that most people do not enjoy traveling and only do so to fulfill necessary activities. %On the other hand, the reward values for subsequent travel steps are higher than for the initial step, suggesting that people may care more about travel frequencies than travel duration. 
Another interesting observation is that the instant reward for the same activity can change as the stay duration increases. Generally, the reward first increases but starts to decrease once the stay duration exceeds a certain threshold, indicating that people have a preferred stay duration for different activities.

\begin{figure}[!ht]
  \centering
  \includegraphics[width=\textwidth]{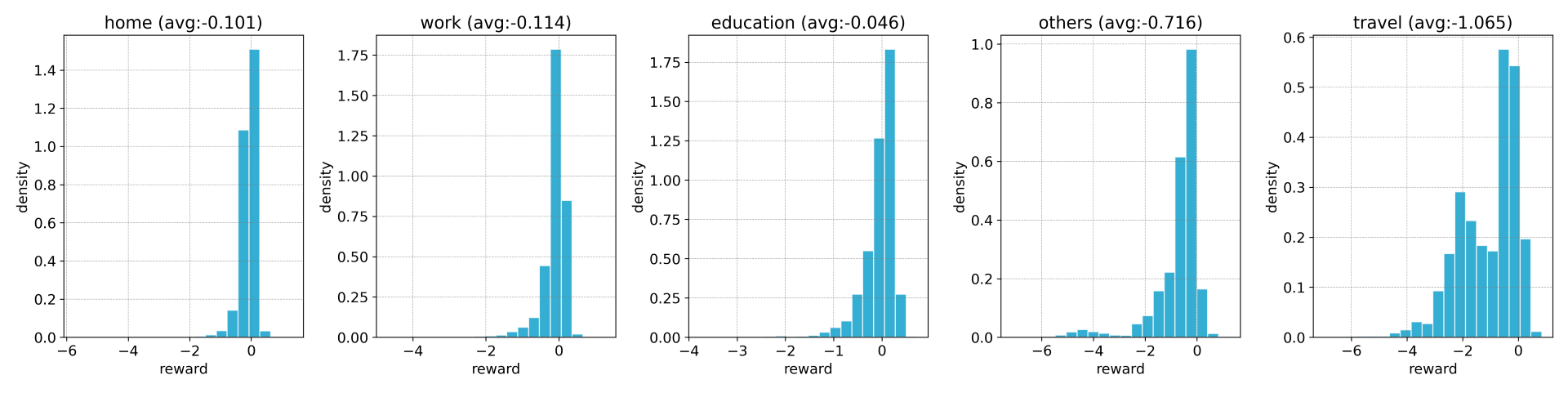}
  \caption{Reward distribution of different activities}\label{fig:reward_distribution}
\end{figure}

Figure~\ref{fig:reward_distribution} depicts the distribution of reward values across various activity types. The mean reward values for home, work, and education activities are higher compared to other activities. This is likely because home, work, and education are regular, expected parts of daily life, whereas activities like shopping or medical visits arise from irregular personal or household needs that are less predictable. On the other hand, travel has the lowest average reward, which aligns with previous findings that it is generally perceived as a necessary but often unpleasant activity. The variance in reward values is also noteworthy. The distribution of rewards for other activities is more dispersed compared to home, work, and education. This is reasonable since ``other'' activities encompass a wide range of possible categories, such as shopping, pick-up drop-off, and medical visits, each with different reward values. The reward values for travel show two peaks. This can be explained by the extremely low rewards associated with the initial transition into travel, while subsequent travel steps are associated with higher reward values, as illustrated in Figure~\ref{fig:reward_examples}.
%Home provides safety, comfort, and personal space; work meets financial needs and provides a sense of accomplishment; and education is often viewed as an investment in personal development and future well-being. 

\begin{figure}[!ht]
  \centering
  \includegraphics[width=\textwidth]{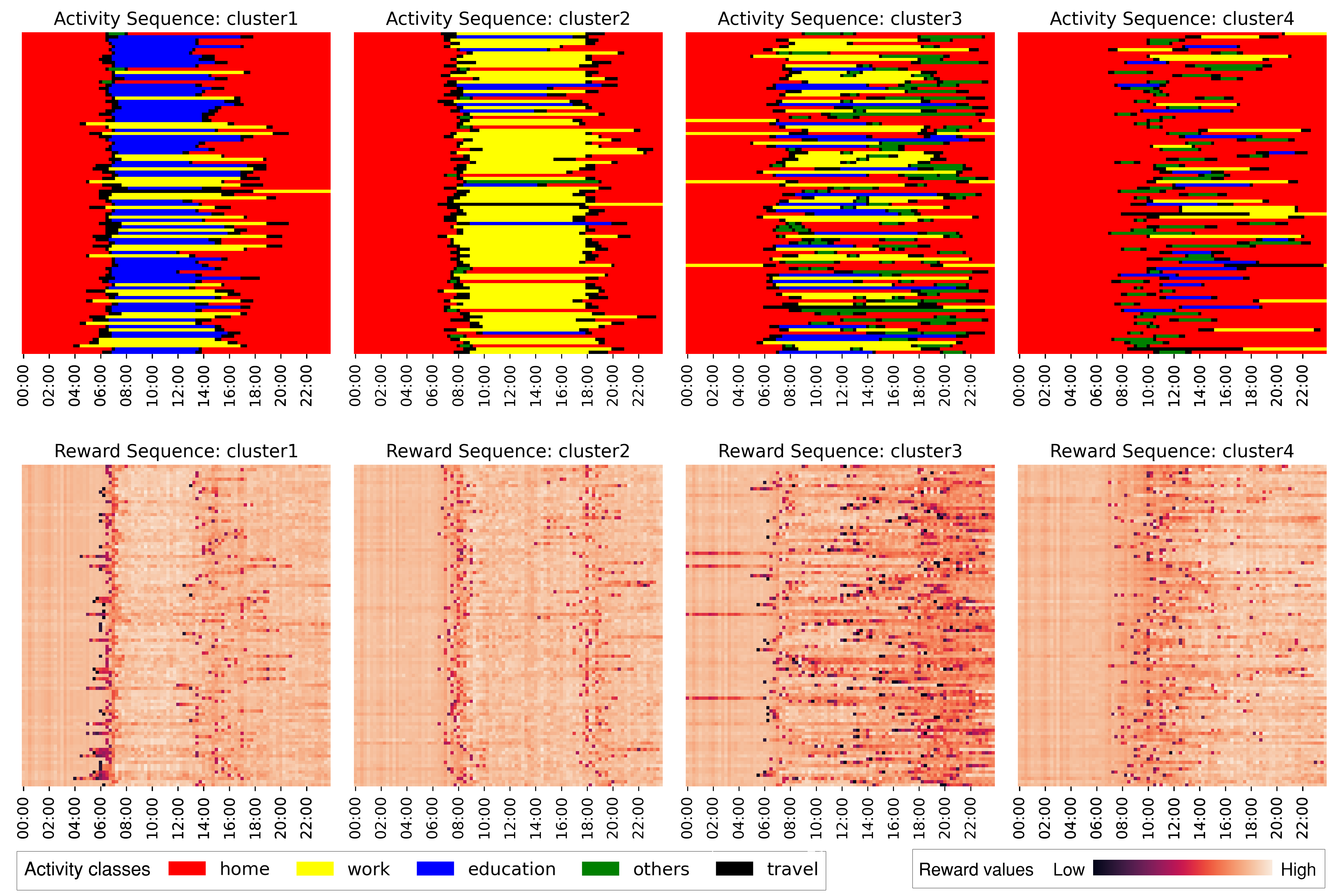}
  \caption{Reward distribution of different activities}\label{fig:reward_clustering}
\end{figure}

Reward values at different timesteps are correlated with each other. To understand these sequence-level patterns, we use a k-means clustering technique, as described in Section~\ref{method:reward_interpret}, to group reward sequences into different types. The cluster size is set to 4, which is the optimal group size determined using the elbow method. Figure~\ref{fig:reward_clustering} shows the clustering results of different reward sequences and their corresponding activity patterns. %The clustering results clearly indicate there are different types of activity planners with distinct reward and travel patterns. 
Cluster 1 and 2 correspond to users with regular travel patterns at similar times while cluster 3 and 4 correspond to users with less regular and predictable travel patterns. This in some way matches with the well-established explorer and returner theory \citep{pappalardo2015returners}, which distinguishes travelers as either having regular, repetitive movements or exploring new and diverse activities with less predictable movement patterns. %Each cluster reveals distinct reward patterns and their correlations to various activity patterns. In Cluster 1, the reward sequence typically shows two very low reward periods from 6:00-8:00 and 14:00-18:00, with higher rewards in between, reflecting school attendance without additional activities. Cluster 2 displays a similar pattern but with extended high rewards during the day, corresponding to work schedules, with low rewards from 6:00-8:00 and 16:00-22:00. Cluster 3 features lower rewards in the afternoon and evening, indicating activities after school or work. Cluster 4 shows a significant low reward period from 8:00-12:00, likely corresponding to other activities taking place in the morning.
%In Cluster 1, the reward sequence typically exhibits two extremely low reward values during 6:00-8:00 and 14:00-18:00, with generally higher reward values in between, compared to early morning and late afternoon. This reward pattern corresponds to activity sequences where individuals go to school during the day without engaging in additional activities. Cluster 2 shows a similar pattern, although the daytime period with higher reward values is longer, with low reward values appearing during 6:00-8:00 and 16:00-22:00. This cluster corresponds to people who go to work during the day. In Cluster 3, activities in the afternoon and evening are associated with lower reward values, corresponding to other activities that occur after school or work in the afternoon or night. In Cluster 4, there is a noticeable low reward period from 8:00-12:00, likely corresponding to other activities taking place in the morning. 

\begin{figure}[!ht]
  \centering
  \includegraphics[width=\textwidth]{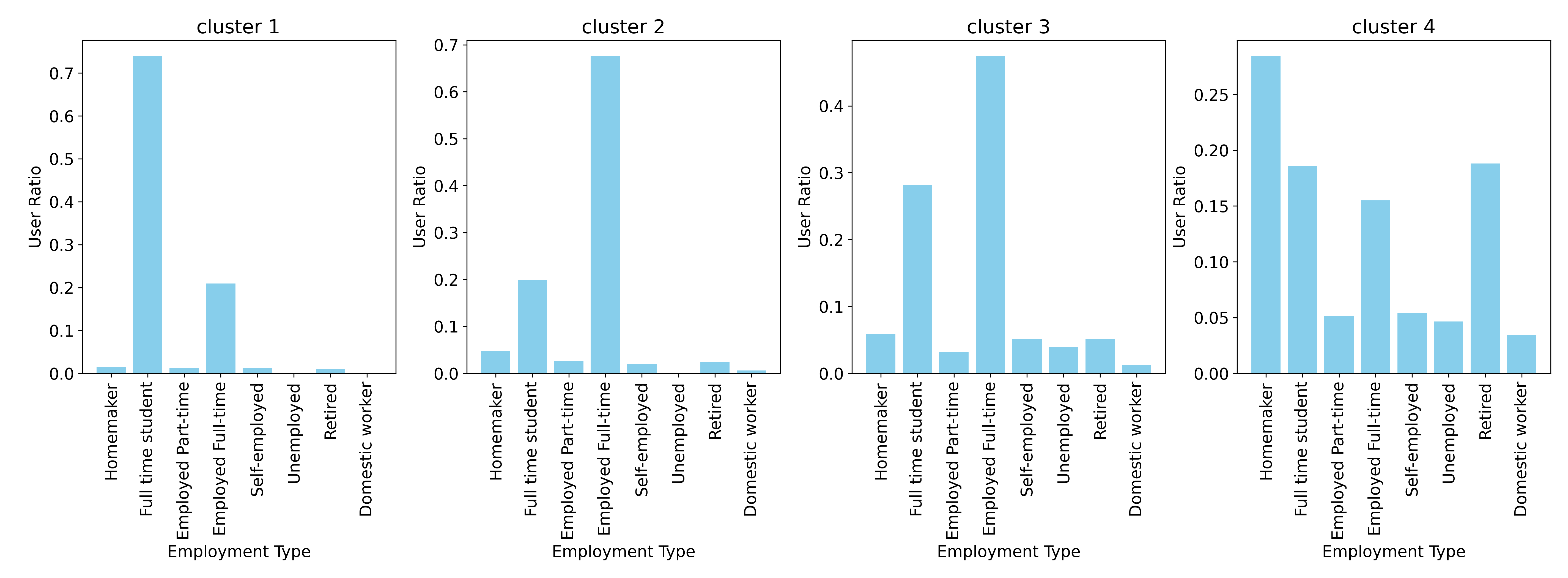}
  \caption{Reward distribution of different activities}\label{fig:reward_employment}
\end{figure}

To understand the individual variance between different groups, we analyze the employment type composition as shown in Figure~\ref{fig:reward_employment}. Interestingly, each group is dominated by different employment types. Cluster 1 is primarily composed of full-time students, exhibiting representative school patterns. Cluster 2 is dominated by full-time employees, which aligns with typical commuting patterns. Cluster 3 includes a mix of full-time students and full-time employees, corresponding to individuals with regular work or school schedules who also engage in additional activities in the afternoon or evening. Cluster 4 is dominated by homemakers and retired individuals, explaining why they perform other activities during the morning hours. This demonstrates the capability of reward values to reflect and differentiate between various activity patterns and individual behaviors. 

\subsubsection{Long-term return}\label{res:long_return}

\begin{figure}[!ht]
  \centering
  \includegraphics[width=\textwidth]{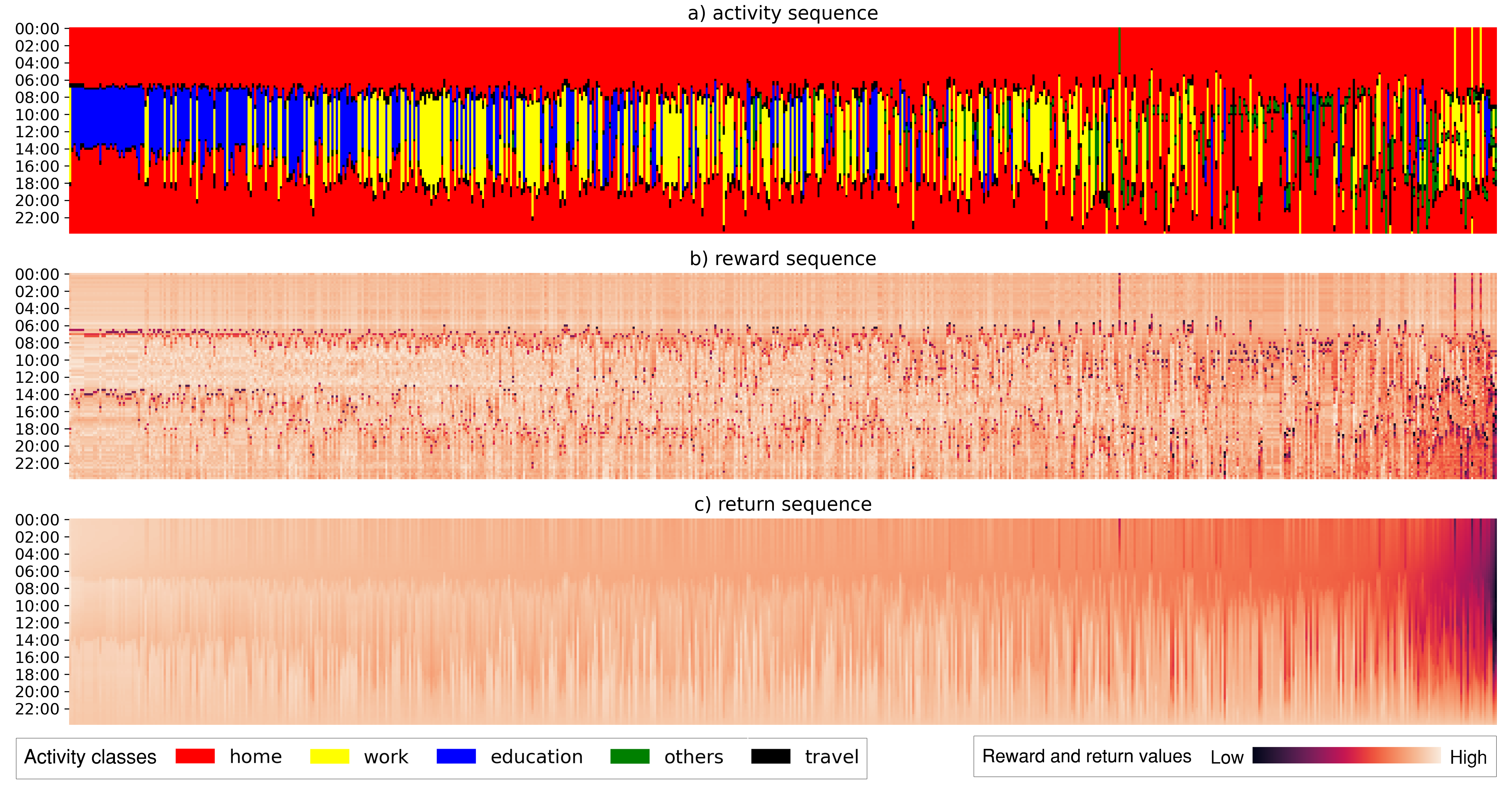}
  \caption{Activity, reward, and return sequences sorted by daily return values (high to low from left to right)}\label{fig:sorted_return_examples}
\end{figure}

Based on the reward values at each step, we can estimate the long-term return of the activity sequence in a day, which compresses the impact of different activities at different time steps into a single value. The long-term return in a day can in some way indicate the overall utility an individual can get from his or her daily activity planning. By sorting the activity sequence based on their corresponding return value, we can then have a sense of what types of activity planning are more optimal than others, at least based on the understanding of the DIRL model. Figure~\ref{fig:sorted_return_examples} shows 500 activity sequences sorted by their associated return values. Activity sequences on the left side are associated with higher returns. The results clearly demonstrate a change from regular to irregular activity patterns, indicating that the model assigns higher return values for individuals with regular activity patterns. Activity sequences with top return values are school patterns, followed by regular commuting patterns. This corresponds to the results in Figure~\ref{fig:reward_distribution} that the DIRL model assigns a higher reward value to education than to work. This can be because people have a better sense of accompliment and get better long-term investiment when experiencing education. 

\begin{figure}[!ht]
  \centering
  \includegraphics[width=\textwidth]{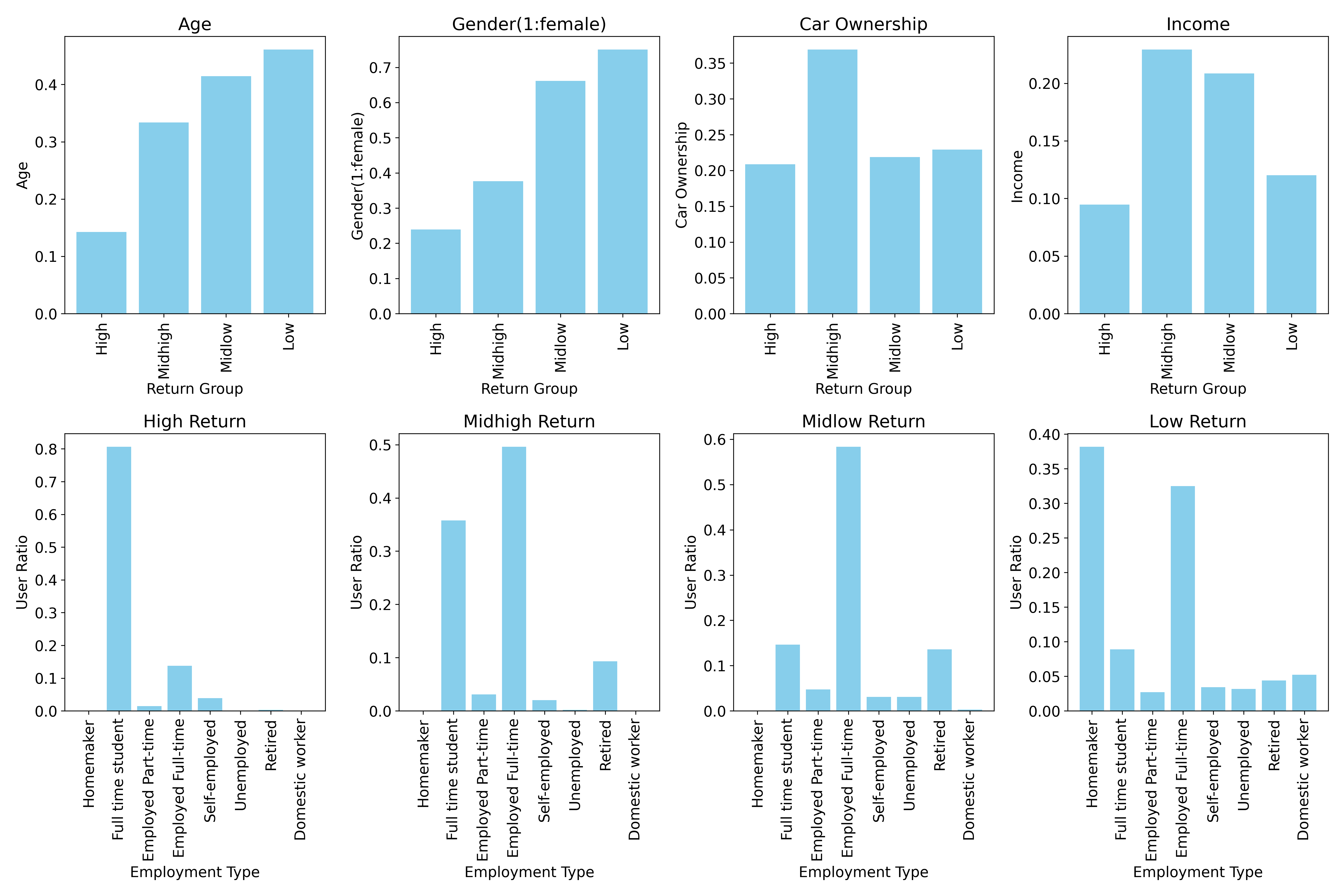}
  \caption{Socio-demographic features of users with different daily return levels}\label{fig:return_demographics}
\end{figure}

To explain the individual variance in return values associated with daily activity sequences, we divided trajectories into four groups based on return quantiles. %The top 0-25\% of activity sequences are associated with high returns, the 25-50\% with mid-high returns, the 50-75\% with mid-low returns, and the 75-100\% with low returns. 
Figure~\ref{fig:return_demographics} shows the socio-demographic features of these four groups. Interestingly, the employment type compositions associated with user groups having different return levels closely align with the types of activity planners identified through the clustering technique discussed in the previous section. The top return group consists mostly of students. The mid-high return group is a mix of full-time employees and students. The mid-low return group is mainly full-time employees, while the low return group is predominantly homemakers. This distribution indicates that return values can reflect differences in employment types among individuals. Regarding age, groups with lower returns tend to include older individuals. This is likely because students, who receive higher returns, are typically younger, while homemakers and retired individuals, who receive lower returns, are generally older. There are also gender differences in return values: in the top return group, only 20\% of individuals are female, whereas in the bottom return group, more than 70\% are female. This suggests that females are typically associated with lower daily returns, possibly due to taking on more home duties that do not directly generate socially recognized value. The relationship between income and return groups forms a ``$\cap$" shape: the top return group has the lowest average income, which is reasonable as it mainly consists of full-time students with no income. Conversely, from the mid-high to low return groups, as the average income decreases, the return on their daily activities also decreases. This indicates that the return values derived from activity sequences have the potential to reflect individual income levels.

\section{Conclusion and Discussion}

\noindent This study presents an interpretable deep inverse reinforcement learning (DIRL) framework designed to uncover sequential activity-travel behavior patterns. Unlike previous research, which emphasizes the prediction ability of DIRL, our focus is on explaining the underlying behavioral mechanism through enhanced model interpretability. We start by adapting an adversarial IRL framework to model sequential behavior patterns, learning a policy function to replicate human behaviors and a reward function to infer human preferences. To extract explainable insights from the policy function, we introduce a knowledge distillation method. Specifically, we train a surrogate Multinomial Logit (MNL) model using the soft labels predicted by the trained policy network, allowing us to extract knowledge from the parameters of MNL. The learned reward function generates a reward value for each state-action pair, providing a quantifiable measure of immediate human preferences at each step. We apply clustering techniques to these reward values to distinguish different types of activity planners. Additionally, we calculate a long-term return value for each behavioral sequence, demonstrating the potential to infer the overall utility an individual gains through a sequential decision-making process.

We analyze travel survey data from Singapore using our proposed frameworks. The results reveal differences in activity and travel behavior across various employment types. Part-time and self-employed workers travel less frequently than full-time workers, indicating that work modes significantly influence travel tendencies. Homemakers and retirees engage in more other activities beyond school and work, while unemployed individuals, despite having flexible schedules, tend to stay home rather than move around. These patterns suggest that human movement can be inferred from employment types, highlighting the link between mobility and socio-economics. Policymakers could potentially influence mobility through economic adjustments that alter employment compositions. Behavioral differences are also evident among socio-demographic groups. Older individuals have higher travel probabilities, likely due to social and household duties. Males are more likely to travel to school and work, while females tend to stay home or engage in other activities, raising concerns about gender inequality in mobility. By clustering reward sequences, we identify four types of activity planners: (1) regular workers commuting between home and work, (2) regular schoolers commuting between home and school, (3) less regular workers or schoolers who engage in other activities before returning home, and (4) homemakers or retirees who conduct morning activities. This distinction among activity planners can potentially serve as a foundation for urban and transport planners to understand various user reactions when designing policy interventions. We further quantify the preference of sequential decision-making processes with a single return value. Regular travel patterns, typical of students and workers, yield higher returns, whereas irregular patterns, typical of homemakers, yield lower returns. Individuals with lower activity returns are often female, older, and have lower incomes, indicating the need for greater attention to minority groups whose activities may lack social recognition.
% , suggesting which activity patterns are more preferential

This study can be further improved in several ways. Firstly, although our current focus is on analyzing activity and travel behavior, the proposed interpretable DIRL framework is general and can be adapted to various sequential decision-making processes. These include route choice modeling in transportation \citep{zhao2023deep}, trading strategies and risk preferences for investment \citep{yang2018investor}, and patient treatment strategies in medical domains \citep{babes2011apprenticeship}. Future studies can extend this framework to other scenarios as relevant data becomes available. Secondly, we currently model daily activity and travel patterns due to data limitations, which may not fully capture travel demand management strategies across different days of the week. Future research could explore weekly or monthly activity planning strategies using data with longer time coverage, such as mobile phone data. Thirdly, our current modeling process focuses on activity type choice and scheduling behavior. Future research could incorporate more dimensions, such as destination choice and mode choice. A significant challenge in modeling all these elements within an DIRL framework is the large space of action choice candidates. To manage the complexity of modeling these elements, a potential solution is to develop a hierarchical DIRL framework \citep{chen2023hierarchical}, which breaks down the decision-making process into a hierarchy of simpler sub-tasks. Lastly, the extent to which the utility values derived from IRL align with actual human perceptions remains largely unexplored. In this study, we offer some evidence that activity sequences with lower return values tend to be associated with individuals of lower income, indicating potential correlations. However, the real-world implications of the utility values inferred by the model still require further exploration.

%\section*{Acknowledgements}
%This research is supported by National Natural Science Foundation of China (NSFC 42201502).

%% The Appendices part is started with the command \appendix;
%% appendix sections are then done as normal sections
% \appendix

% \section{Methods for Missing Pattern Generation}\label{missing_generation}

% To Add

%% \section{}
%% \label{}

%% References
%%
%% Following citation commands can be used in the body text:
%% Usage of \cite is as follows:
%%   \cite{key}          ==>>  [#]
%%   \cite[chap. 2]{key} ==>>  [#, chap. 2]
%%   \citet{key}         ==>>  Author [#]

%% References with bibTeX database:
%\section*{References}
\bibliographystyle{model5-names2}\biboptions{authoryear}
\bibliography{main}

%% Authors are advised to submit their bibtex database files. They are
%% requested to list a bibtex style file in the manuscript if they do
%% not want to use model1-num-names.bst.

%% References without bibTeX database:

% \begin{thebibliography}{00}

%% \bibitem must have the following form:
%%   \bibitem{key}...
%%

% \bibitem{}

% \end{thebibliography}

\end{document}